\documentclass[11pt, a4paper, logo, copyright]{googlecloud}

\pdftrailerid{redacted}

\makeatletter
\renewcommand\bibentry[1]{\nocitep{#1}{\frenchspacing\@nameuse{BR@r@#1\@extra@b@citeb}}}
\makeatother

\usepackage{kantlipsum, lipsum}
\usepackage{dsfont}
\usepackage[authoryear, sort&compress, round]{natbib}

\usepackage[utf8]{inputenc} %
\usepackage[T1]{fontenc}    %
\usepackage{hyperref}       %
\hypersetup{
  colorlinks,
  citecolor=blue,
  linkcolor=red,
  urlcolor=blue}
\usepackage{booktabs}       %
\usepackage{nicefrac}       %
\usepackage{microtype}      %
\usepackage{wrapfig} %
\usepackage{soul} %
\usepackage{tikz, adjustbox}
\usepackage{arydshln}
\usepackage[capitalize,noabbrev]{cleveref}
\usepackage{fancyvrb}

\usepackage{inconsolata}
\usepackage[skins,breakable,most]{tcolorbox}

\usepackage{caption}
\usepackage{subcaption}
\usepackage{enumitem}
\usepackage{flushend}
\usepackage{balance}
\usepackage{lineno}
\usepackage{amsmath,amssymb,amsfonts}
\usepackage{algorithm}
\usepackage{algpseudocode}
\usepackage{graphicx}
\usepackage{textcomp}
\usepackage{xcolor}
\usepackage{colortbl}
\usepackage{amsthm}
\usepackage{url}
\usepackage{float}
\usepackage{multirow}
\usepackage{multicol}
\usepackage{color}
\usepackage{bm}
\usepackage{bbm}

\usepackage{pifont}

\usepackage{array}
\newcolumntype{L}[1]{>{\raggedright\let\newline\\\arraybackslash\hspace{0pt}}m{#1}}
\newcolumntype{C}[1]{>{\centering\let\newline  \\\arraybackslash\hspace{0pt}}m{#1}}
\newcolumntype{R}[1]{>{\raggedleft\let\newline \\\arraybackslash\hspace{0pt}}m{#1}}

\usepackage[utf8]{inputenc} 
\usepackage[T1]{fontenc}    
\usepackage{amsfonts}       
\usepackage{comment} 
\usepackage{mdframed}
\usepackage{fontawesome}
\usepackage{csquotes}
\usepackage{makecell}
\definecolor{beigecolor}{RGB}{253, 244, 204} 
\definecolor{greencolor}{RGB}{228, 242, 217} 
\definecolor{bluecolor}{RGB}{66, 133, 244} 
\definecolor{orgcolor}{RGB}{255, 140, 15} 

\definecolor{redcolor}{RGB}{234, 67, 53} 

\definecolor{ggreen}{RGB}{52, 168, 83}

\definecolor{gyellow}{RGB}{251, 188, 5}

\lstdefinestyle{mystyle}{
    backgroundcolor=\color{backcolour},   
    commentstyle=\color{codegreen},
    keywordstyle=\color{magenta},
    numberstyle=\tiny\color{codegray},
    stringstyle=\color{codepurple},
    basicstyle=\ttfamily\scriptsize,
    breakatwhitespace=false,         
    breaklines=true,                 
    captionpos=b,                    
    keepspaces=true,                 
    numbers=left,                    
    numbersep=5pt,                  
    showspaces=false,                
    showstringspaces=false,
    showtabs=false,                  
    tabsize=2,
    frame=none,
    aboveskip=1pt,
    belowskip=1pt,
}
\lstdefinestyle{plainins}{
    backgroundcolor=\color{white},   
    commentstyle=\color{codegreen},
    keywordstyle=\color{magenta},
    numberstyle=\tiny\color{codegray},
    stringstyle=\color{codepurple},
    basicstyle=\ttfamily\scriptsize,
    breakatwhitespace=false,         
    breaklines=true,                 
    captionpos=b,                    
    keepspaces=true,                 
    numbers=none,                    
    numbersep=5pt,                  
    showspaces=false,                
    showstringspaces=false,
    showtabs=false,                  
    tabsize=2,
    aboveskip=0pt,
    belowskip=0pt,
    frame=single
}
\lstdefinestyle{plainexam}{
    backgroundcolor=\color[HTML]{FFFCF3},   
    commentstyle=\color{codegreen},
    keywordstyle=\color{magenta},
    numberstyle=\tiny\color{codegray},
    stringstyle=\color{codepurple},
    basicstyle=\ttfamily\scriptsize,
    breakatwhitespace=false,         
    breaklines=true,                 
    captionpos=b,                    
    keepspaces=true,                 
    numbers=none,                    
    numbersep=5pt,                  
    showspaces=false,                
    showstringspaces=false,
    showtabs=false,                  
    tabsize=2,
    aboveskip=0pt,
    belowskip=0pt
}

\title{\textit{In Prospect and Retrospect}: Reflective Memory Management for Long-term Personalized Dialogue Agents}

\correspondingauthor{ztan36@asu.edu, \{junyann, chenyulee\}@google.com}

\renewcommand{\today}{}

\author[1 *]{Zhen Tan}
\author[2]{Jun Yan}
\author[2]{I-Hung Hsu}
\author[2]{Rujun Han}
\author[2]{Zifeng Wang}
\author[2]{Long T. Le}
\author[2]{Yiwen Song}
\author[2]{Yanfei Chen}
\author[2]{Hamid Palangi}
\author[2]{George Lee}
\author[3]{Anand Iyer}
\author[4]{Tianlong Chen}
\author[1]{Huan Liu}
\author[2]{Chen-Yu Lee}
\author[2]{Tomas Pfister}

\affil[1]{Arizona State University}
\affil[2]{Google Cloud AI Research}
\affil[3]{Google Cloud AI}
\affil[4]{UNC Chapel Hill}

\begin{abstract}
Large Language Models (LLMs) have made significant progress in open-ended dialogue, yet their inability to retain and retrieve relevant information from long-term interactions limits their effectiveness in applications requiring sustained personalization. External memory mechanisms have been proposed to address this limitation, enabling LLMs to maintain conversational continuity. However, existing approaches struggle with two key challenges. First, rigid memory granularity fails to capture the natural semantic structure of conversations, leading to fragmented and incomplete representations. Second, fixed retrieval mechanisms cannot adapt to diverse dialogue contexts and user interaction patterns.
In this work, we propose \textbf{Reflective Memory Management} (\textbf{RMM}), a novel mechanism for long-term dialogue agents, integrating forward- and backward-looking reflections: (1) \textbf{Prospective Reflection}, which dynamically summarizes interactions across granularities—utterances, turns, and sessions—into a personalized memory bank for effective future retrieval, and (2) \textbf{Retrospective Reflection}, which iteratively refines the retrieval in an online reinforcement learning (RL) manner based on LLMs' cited evidence. Experiments show that RMM demonstrates consistent improvement across various metrics and benchmarks. For example, RMM shows more than 10$\%$ accuracy improvement over the baseline without memory management on the \texttt{LongMemEval} dataset.
\end{abstract}

\begin{document}

\maketitle

\section{Introduction}

\begin{wrapfigure}{r}{0.5\textwidth}
    \centering
    \includegraphics[width=1\linewidth]{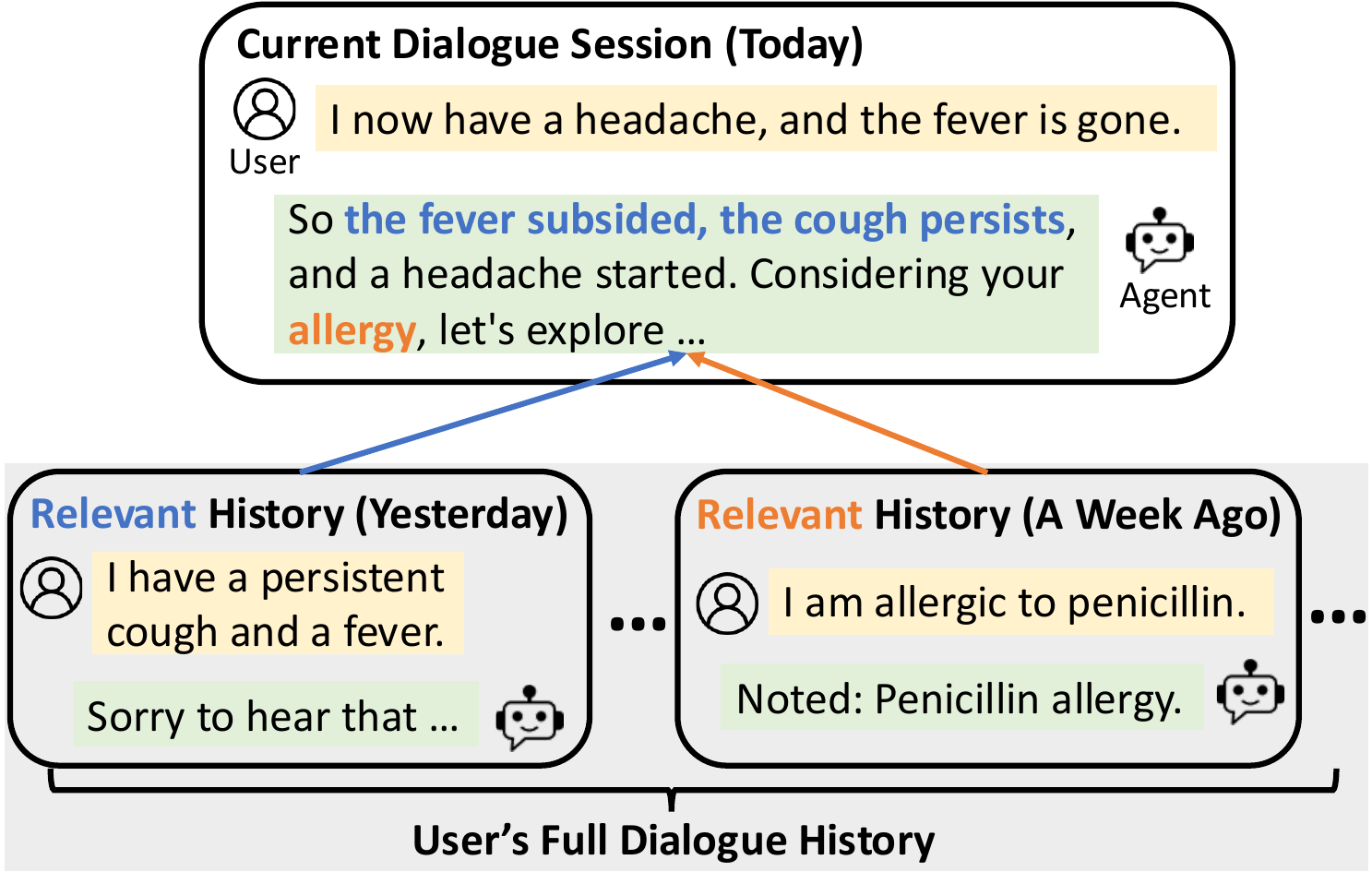}
    \caption{An illustration of a personalized healthcare agent. Key information about a user's allergy and previous symptoms mentioned in the past sessions is needed to provide a more informed response in the current session.}
    \label{fig:teaser}
\end{wrapfigure}

Large Language Models (LLMs) have demonstrated remarkable capabilities in engaging in open-ended dialogue~\citep{lee2023prompted,mendoncca2024benchmarking}, yet their inherent statelessness poses a significant challenge for maintaining coherent, personalized conversations over time~\citep{chen2024large,li2024personal,tseng2024two}, which are crucial across various real-world applications (\textit{e.g.}, customer service~\citep{kolasani2023optimizing}, virtual assistants~\citep{guan2023intelligent}, and education platforms~\citep{zhang2024simulating,wen2024ai}). 
As illustrated in Figure~\ref{fig:teaser}, effective personalization requires not only understanding the immediate context but also recalling relevant information from the user's previous interactions~\citep{williams1981process,whittaker2002managing,dong2024can}.
The limitations with current LLMs to naturally retain and recall information from past interactions beyond their context windows sparked the development of external memory mechanisms for LLMs~\citep{zhang2024personalization,li2024hello,kim2024theanine}.
These memory systems serve as crucial components in personalized dialogue agents, enabling them to maintain consistent personality traits, remember user preferences, and build upon previous interactions.

While external memory mechanisms represent a significant step towards enabling persistent dialogue, current approaches suffer from two critical limitations. Firstly, existing systems digest information at a pre-defined granularity, such as turn, session, or time interval boundaries, which may not align with the inherent semantic structure of the conversation (e.g., topic shifts). This rigid approach can lead to fragmented or incomplete memory representations, hindering the LLM's ability to retrieve, utilize, and update relevant information effectively~\citep{wu2024longmemeval,pansecom}. Secondly, these systems rely on fixed retrievers~\citep{zhong2024memorybank,li2024hello}, which struggle to adapt to the diverse retrieval demands of varying dialogue domains and individual user interaction patterns. Moreover, the expense associated with collecting labeled data for training personalized retrievers presents a substantial barrier to widespread adoption and scalability.

To address these limitations, we propose a novel \textbf{Reflective Memory Management} (\textbf{RMM}) mechanism to provide a more adaptable and granular approach to long-term dialogue memory. Our framework incorporates two key innovations. \textbf{Prospective Reflection} tackles the issue of fixed granularity by summarizing dialogue histories into decomposed topics, effectively integrating fragmented conversational segments into cohesive 
memory structures. This approach optimizes memory organization for future retrieval, allowing the LLM to access relevant information more effectively regardless of the original turn or session boundaries. Complementing this, \textbf{Retrospective Reflection} addresses the challenge of fixed retrievers by leveraging unsupervised attribution signals generated during the LLM's response generation to reflect on past retrieval. This allows for online refinement of the retriever as the conversation progresses, enabling the system to adapt to diverse dialogue domains and individual user interaction patterns without the need for costly labeled data.

By integrating these two reflective mechanisms, our approach enables LLMs to maintain a more nuanced and adaptable memory, leading to more coherent, personalized, and engaging dialogues.
Experiments on \texttt{MSC} and \texttt{LongMemEval} benchmarks show that RMM achieves more than 5\% improvement over the strongest baseline across memory retrieval and response generation metrics.

Our contributions are as follows: (1)~We propose RMM as a novel memory management mechanism that employs topic-based memory management optimized for future retrieval and leverages attribution signal to reflect on past retrieval for unsupervised online retrieval refinement. (2) We conduct extensive experiments on two long-term personalized dialogue benchmarks to demonstrate the effectiveness of RMM over strong baselines. (3)~We perform detailed analysis on the impact of various design choices to pinpoint the limitations of existing memory management mechanisms with fixed granularity and retrievers, shedding light on the room for future improvement.

\section{Related Work}

\noindent\textbf{Long-term Conversations for LLMs.}
LLMs have demonstrated the ability to engage in extended, coherent dialogues, yet maintaining context and consistency over long-term interactions remains a challenge. 
\citet{maharana2024evaluating} introduced the LoCoMo dataset to assess LLMs' performance in sustained dialogues, showing their struggles with long-range temporal and causal understanding. Existing solutions can be broadly categorized into two approaches:
(1) Architectural modifications, such as enhancing attention mechanisms~\citep{liu2023ring,zhang2024spar}, optimizing KV caches~\citep{li2024scbench,liu2025chunkkv}, and refining position embeddings~\citep{zhao2023length,zheng2025dape}. These methods require white-box access to model internals, making them infeasible for proprietary or API-based LLMs.
(2) Summarization-based methods, which condense long contexts into structured events or topics for direct conditioning or retrieval
~\citep{lu2023memochat,wang2023recursively,jiang2024retrieve,li2024alr}.
RMM falls into this category but explicitly addresses the issue of fragmented topics arising from fixed granularity and incorporates retrospective reflection to refine the retrieval process, encouraging more coherent and contextual responses.

\noindent\textbf{Memory-based Personalized Dialogue Agents.} The development of memory-based personalized dialogue agents has further enhanced long-term interactions by enabling systems to retain and utilize information from past conversations~\citep{bae2022keep}. 
Traditional methods \cite{weizenbaum1966eliza,walker1997paradise} laid the groundwork for understanding how systems can model user preferences, intentions, and behaviors across sessions, often using handcrafted rules, heuristics, or symbolic representations.
Early approaches, such as CoMemNN~\citep{pei2021cooperative}, introduce mechanisms to incrementally enrich user profiles during dialogues. However, collecting substantial annotations for training a personalized system for long-term use is hard~\citep{tseng2024two}.
Recent advancements focus on integrating LLMs with memory modules~\citep{packer2024memgptllmsoperatingsystems,chhikara2025mem0buildingproductionreadyai,wang2024agentworkflowmemory,xu2025mem,rasmussen2025zep}. For instance, the LD-Agent framework~\citep{li2024hello} employs long-, short-term memory banks to manage conversational history for retrieval. MemoryBank~\citep{zhong2024memorybank} incorporates a memory updating mechanism inspired by the Ebbinghaus Forgetting Curve, enabling models to retrieve relevant memories considering recency. Theanine~\citep{kim2024theanine} introduces timeline-based retrieval and utilizes an additional LLM for refinement. 
These methods typically deploy fixed retrievers with a pre-defined granularity. In contrast, the proposed RMM approach facilitates adaptive retrieval with a revised retrieval granularity.

\section{Problem Formulation}\label{sec:prob}
We consider the task of building a personalized dialogue agent in a \textbf{multi-session} conversational setting. In this setting, an agent interacts with a user across multiple distinct sessions.
A \textit{session} represents a distinct interaction period, often delimited by user inactivity, explicit user confirmation of conversation completion, or the initiation of a new dialogue thread.
Within each session, the conversation unfolds as a sequence of turns, where a \textit{turn} consists of a user query and the agent's corresponding response.
The agent is equipped with an external memory, serving as the sole repository for information gathered from previous sessions.
The agent's objective is to generate contextually relevant and personalized responses to user queries, leveraging both the immediate conversational context within the current session and the relevant information retrieved from the memory.

This task presents two key challenges: first, the agent must proactively identify and store salient information from each session, anticipating \textit{future} retrieval needs. Second, the agent must accurately retrieve relevant \textit{past} information from the memory, as incorporating irrelevant context can distract the LLM and degrade response quality \citep{shi2023large,liu2024lost}. Effectively managing this balance between comprehensive storage and precise retrieval is critical for achieving personalized and coherent multi-session dialogues.

\section{Framework Overview}

To tackle the challenges, we introduce Reflective Memory Management (RMM), a novel framework that integrates two mechanisms. Prospective Reflection proactively decomposes dialogue history into topic-based memory representations, optimizing for future retrieval, while Retrospective Reflection dynamically refines the retrieval mechanism through online feedback signals generated during response generation.
They together improve the quality of the retrieved memories, contributing to effective personalization.

\begin{wrapfigure}[18]{r}{0.5\textwidth}
\begin{minipage}{0.5\textwidth}
\vspace{-3mm}
\begin{algorithm}[H]
\small
\captionsetup{font=small}
\caption{Reflective Memory Management (RMM) for Dialogue Agents}
\label{alg:rmm}
\textbf{Input:} query $q$, past messages in current session $S$, memory bank $B$, retriever $f_\theta$, reranker $g_\phi$, $\textsc{LLM}$\\
\textbf{Output:} response $a$, updated $S$, $g_\phi$, $B$
\begin{algorithmic}[1]
\State \textbf{Retrieve:} $\mathcal{M}_K \gets f_\theta(q,B)$
\State \textbf{Rerank:} $\mathcal{M}_M \gets g_\phi(q, \mathcal{M}_K)$, where $\mathcal{M}_M=\{m_i\}_{i=1}^{M}$
\State \texttt{\textsc{// Retrospective Reflection}}
\State \textbf{Generate:} $a, R_M \gets \textsc{LLM}(q, S, \mathcal{M}_M)$ where $R_M=\{r_i\}_{i=1}^M$
\State $g_\phi \gets \textbf{RL\_Update}(g_\phi, R_M)$
\State $S.append((q, a))$
\State \texttt{\textsc{// Prospective Reflection}}
\If{session $S$ ends}
    \State $\mathcal{M} \gets \textbf{ExtractMemory}(S)$
    \For{$m \in \mathcal{M}$}
        \State $B \gets \textbf{UpdateMemory}(B, m)$
    \EndFor
    \State $S \gets []$
\EndIf
\end{algorithmic}
\end{algorithm}
\end{minipage}
\end{wrapfigure}

Our framework comprises four key components. The \textbf{memory bank} stores dialogue history as a collection of memory entries, each represented as a pair (topic summary, raw dialogue), where the ``topic summary'' serves as the search key for retrieving the conversational segment. The \textbf{retriever} identifies relevant memories based on the current user query. To enable lightweight adaptation of the retrieval process, we incorporate a \textbf{reranker}, which refines the retriever's initial output by prioritizing the most pertinent memories. Finally, an \textbf{LLM} synthesizes the relevant memories with the current context to produce a personalized response. Crucially, the LLM also provides feedback signals based on its utilization of retrieved memories, which are used to refine the reranker through Retrospective Reflection. Our complete workflow is detailed in Algorithm~\ref{alg:rmm}.

\section{Prospective Reflection: Topic-Based Memory Organization}

\begin{wrapfigure}{r}{0.5\textwidth}
    \centering
    \includegraphics[width=1\linewidth]{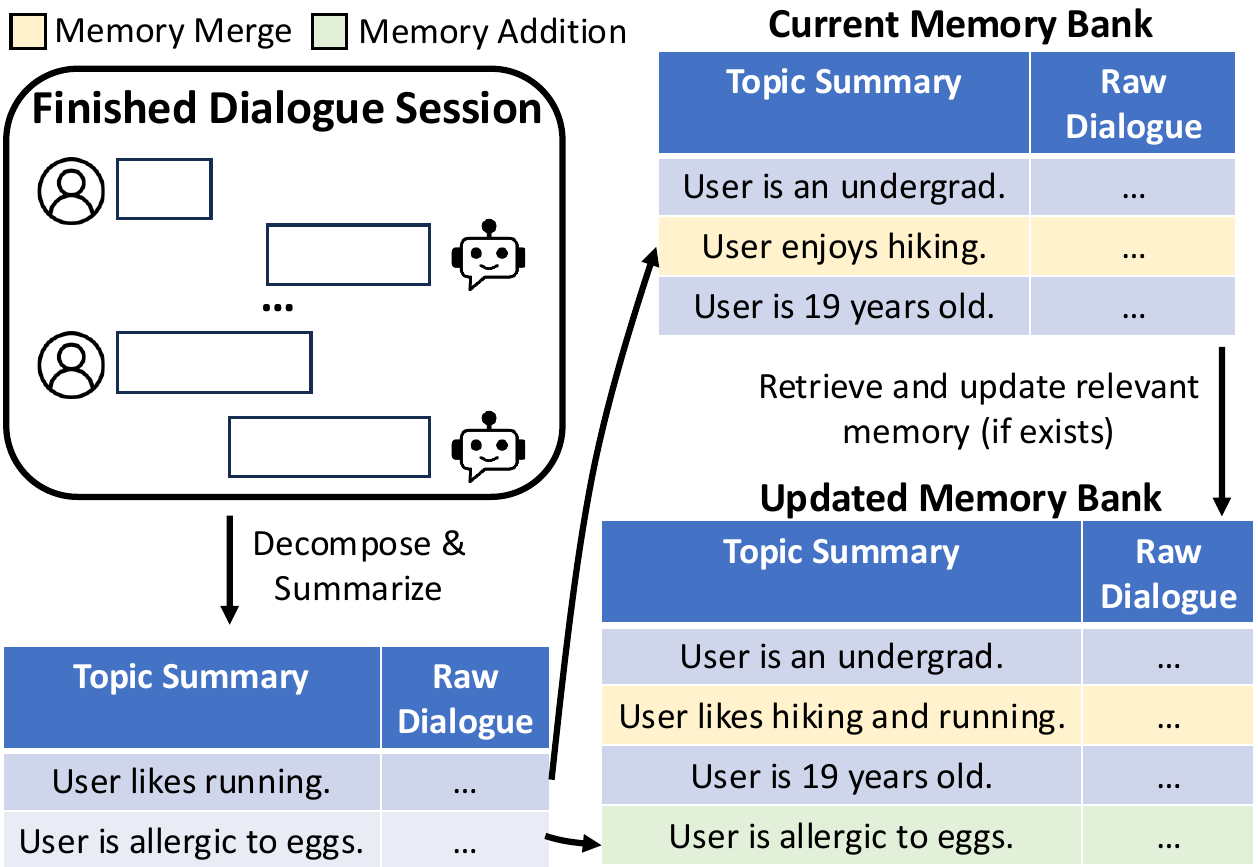}
    \caption{Illustration of \textit{Prospective Reflection}. After each session, the agent decomposes and summarizes the session into specific topics. These newly generated memories are compared with existing memories in the memory bank. Relevant memories are \colorbox{beigecolor}{merged}, while others are directly \colorbox{greencolor}{added}. Prospective reflection ensures efficient organization of personal knowledge for future retrieval.}
    \label{fig:pro}
    \vspace{-4mm}
\end{wrapfigure}

Traditional memory management systems often rely on fixed boundaries, such as session or turn delimiters, to structure dialogue history. However, these pre-defined boundaries may not align with the underlying semantic units of conversation. As a result, critical information may be fragmented across multiple memory entries, hindering effective retrieval. To address this, we introduce Prospective Reflection, a mechanism for organizing memory based on coherent topics, enabling more granular and semantically relevant future retrieval. Here ``topic'' refers to a semantically coherent unit of discussion that may span across one or multiple turns in a session. Each topic is associated with the raw dialogue segment(s) in which it was discussed. These topics can range from fine-grained user intents (e.g., asking about vegan recipes) to broader themes (e.g., travel planning). As illustrated in Figure~\ref{fig:pro}, this process occurs at the conclusion of each session and consists of two key steps: memory extraction and memory update.

First, \textbf{memory extraction} is achieved by using an LLM (prompt in Appendix~\ref{app:p2}) to extract dialogue snippets from the session with their corresponding summaries based on the distinct mentioned topics. Second, \textbf{memory update} involves integrating the extracted topic-based memories into the memory bank. Specifically, for each extracted memory, we retrieve the Top-$K$ most semantically similar memories already present in the memory bank. Subsequently, an LLM (prompt in Appendix~\ref{app:p3}) determines whether the extracted memory should be directly \textbf{added} into the memory bank (e.g., when the extracted memory discusses a new topic) or \textbf{merged} with an existing memory into an updated one (e.g., when the extracted memory provides updated information to a previously discussed topic).

Through Prospective Reflection, the memory bank maintains a coherent and consolidated representation of the evolving dialogue history, organized around meaningful topic structures.

\section{Retrospective Reflection: Retrieval Refinement via LLM Attribution}

\subsection{Reranker Design}

While an off-the-shelf retriever can identify semantically-relevant memories, its performance can degrade across diverse dialogue domains and user interaction patterns.
Instead of resorting to computationally expensive fine-tuning of the retriever, which requires extensive labeled data, we introduce a lightweight reranker to refine the retrieved memory list.
This reranker allows for efficient adaptation to the nuances of specific dialogue domains and user preferences, enabling the system to dynamically adjust its retrieval strategy.

To be specific, the reranker processes the Top-$K$ memory embeddings retrieved by the retriever, refining their relevance with respect to the user query and selecting the Top-$M$ candidates. The whole process includes the following steps.

\noindent\textbf{Embedding Adaptation.} Let \(\mathbf{q}\) represent the embedding of the query and \(\mathbf{m}_i\) represent the embedding of the \(i\)-th memory entry retrieved by retriever. The embeddings are fed into the reranker to be refined via a linear layer with residual connections:
    \begin{equation}
    \mathbf{q}^{\prime} = \mathbf{q} + \mathbf{W}_q \mathbf{q}, \quad \mathbf{m}_i^{\prime} = \mathbf{m}_i + \mathbf{W}_m \mathbf{m}_i,
    \end{equation}
    where \(\mathbf{W}_q\) and \(\mathbf{W}_m\) are linear transformation matrices for the query and memory, respectively.

\noindent\textbf{Stochastic Sampling with Gumbel Trick.} The adapted query embedding \(\mathbf{q}'\) and memory embeddings \(\mathbf{m}_i'\) are adopted to compute relevance scores via dot product: $s_i = \mathbf{q}'^\top \mathbf{m}_i'$. 
To select memory entries based on relevance scores, we employ the Gumbel Trick~\citep{gumbel1948statistical}, which enables stochastic sampling from a discrete probability distribution while preserving gradients, making it particularly useful in reinforcement learning and differentiable ranking tasks~\citep{jang2016categorical}. 
We add Gumbel noise \(g_i\)~\citep{maddison2014sampling} to the relevance scores \(s_i\) for each memory entry:
\begin{equation}
\tilde{s}_i = s_i + g_i, \quad g_i = -\log(-\log(u_i)),
\end{equation}
where \(u_i \sim \text{Uniform}(0, 1)\). The perturbed scores \(\tilde{s}_i\) are then normalized using the softmax function to compute sampling probabilities:
$p_i = \frac{\exp(\tilde{s}_i / \tau)}{\sum_{j=1}^K \exp(\tilde{s}_j / \tau)}$,
where \(\tau > 0\) is the temperature parameter controlling the sharpness of the distribution. Lower \(\tau\) results in more deterministic sampling (approaching the maximum of \(s_i\)), while higher \(\tau\) increases stochasticity, encouraging exploration.

By introducing a reranker, RMM ensures efficient retrieval refinement without modifying the retriever itself, making it adaptable to any pre-trained retrieval model while allowing task-specific optimizations through Reinforcement Learning (RL).

\subsection{LLM Attribution as Rewards}

Obtaining high-quality user-specific labeled data for refining the retrieval process is prohibitively expensive. To overcome this challenge, we propose leveraging the inherent capabilities of the LLM generator itself to provide automated feedback on the quality of retrieved memories. Given the user query with context in the current session, and the retrieved memories, we prompt the LLM (prompt in Appendix~\ref{app:p1}) to generate both the response and the associated citations to each individual memory in the context~\citep{kenthapadi2024grounding}. This design uses a single LLM call for generating response and LLM attribution, reducing computational overhead. Moreover, the citations are generated conditioned on the response, which has been shown to be more effective compared to prior or post-hoc citations \citep{buchmann-etal-2024-attribute}.

\begin{wrapfigure}{r}{0.5\textwidth}
    \centering
    \includegraphics[width=1\linewidth]{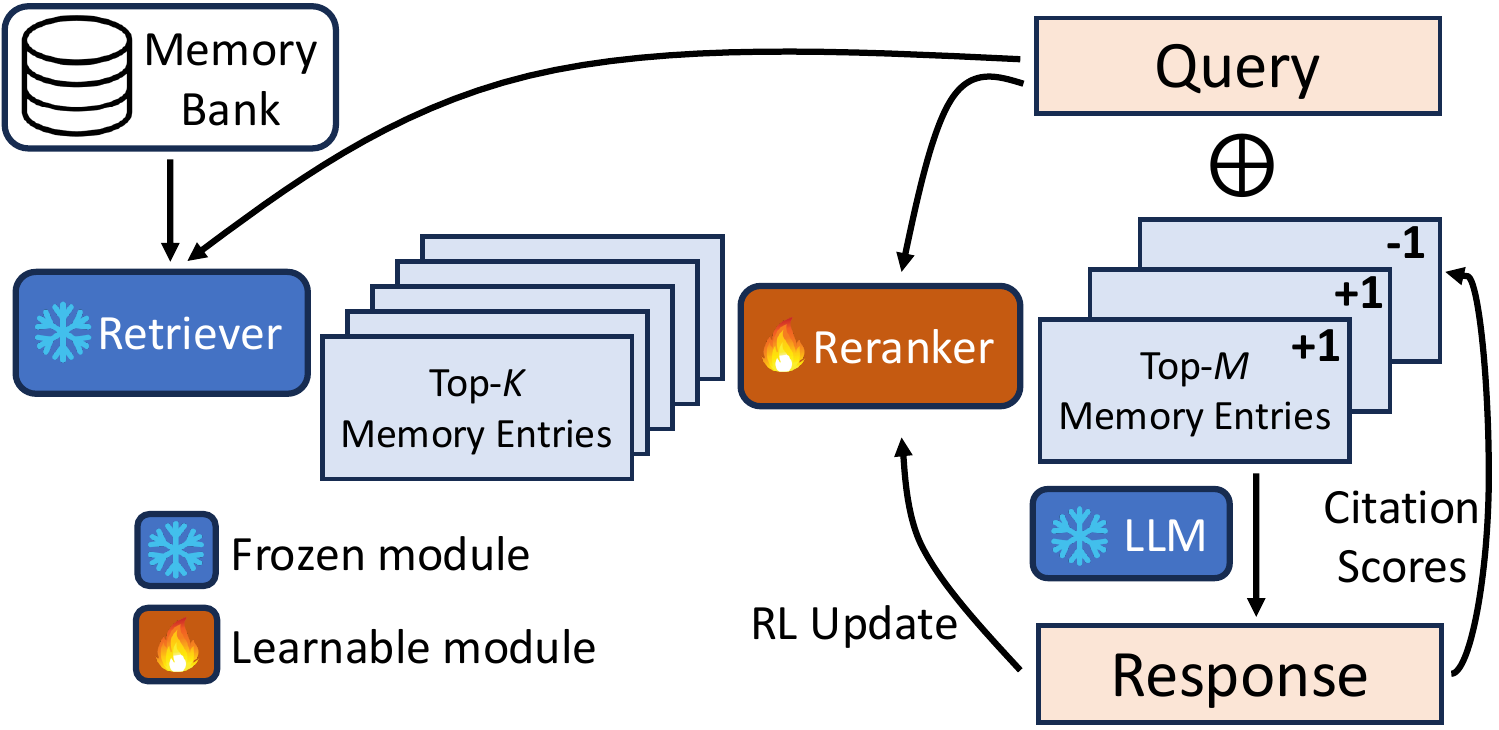}
    \caption{Illustration of \textit{Retrospective Reflection}. The {Retriever} fetches Top-$K$ memory entries from the memory bank, which are refined by the learnable \textit{Reranker} to select the Top-$M$ most relevant entries. These entries are passed to the LLM along with the query to generate the final response. The LLM assigns binary \textit{citation scores} ($+1$ for useful and $-1$ for not useful) to the retrieved memory entries based on their utility in the response. These scores are used as reward signals to update the reranker via an \textit{RL update}, adapting the selection of relevant memory over time.}
    \label{fig:retro}
\end{wrapfigure}

\noindent\textbf{Rewards.} As shown in Figure~\ref{fig:retro}, each retrieved memory entry receives either a positive or negative reward based on its citation in the generated response.
Specifically, we assign a reward of $\mathbf{+1}$ \textbf{(Useful)} if the generator cites the memory in the final response, and $\mathbf{-1}$ \textbf{(Not Useful)} otherwise.
This reward assignment reflects the utility of each memory entry and allows the reranker to learn better retrieval strategies over time, aligning future selections with the generator’s actual usage of retrieved evidence. 
We validate its effectiveness in Section~\ref{sec:cite}.

\subsection{Reranker Update}
The reranker is fine-tuned using the REINFORCE algorithm \citep{williams1992simple} to optimize its relevance predictions based on these binary rewards with the following formulation:
\begin{equation}
\Delta \phi = \eta \cdot (R - b) \cdot \nabla_\phi \log P(\mathcal{M}_M | q, \mathcal{M}_K; \phi),
\end{equation}
where $R$ is the reward ($+1$ or $-1$), $b$ is a baseline value set as a hyperparameter, and $\phi$ denotes the weights of the reranker.

\section{Experimental Setup}
\subsection{Implementation Details}
In our experiments, we use Gemini-1.5-Flash as the generator and evaluate Gemini-1.5-Pro in Section \ref{sec:llm}. We equip RMM with the following dense retrievers with strong semantic representation capabilities and widespread adoption in personalized dialogue systems~\citep{wu2024longmemeval}.
\begin{itemize}[leftmargin=*,itemsep=1pt]
    \item Contriever (\texttt{facebook/contriever}) \citep{izacard2021unsupervised}: A dense retriever optimized for semantic search leveraging contrastive learning.
    \item Stella (\texttt{dunzhang/stella\_en\_1.5B\_v5}) \citep{zhang2024jasper}: A large embedding-based retriever, which is developed based on language models.
    \item GTE (\texttt{Alibaba-NLP/gte-Qwen2-7B-instruct}) \citep{li2023towards}: A retriever designed for instruction-following queries, which is trained across a vast, multilingual text corpus spanning diverse domains.
\end{itemize}
Contriever is used as the default retriever. Following~\citet{wu2024longmemeval}, for experiments without a reranker, the Top-$K$ is 5. Otherwise, the default Top-$K$ is 20 and Top-$M$ is 5. We explore the impact of retrieval parameters in Appendix~\ref{app:topk}.
For LongMemEval, we also consider the results of using an ``Oracle'' retriver which retrieves the ground-truth turns annotated in the dataset with the necessary personal knowledge to respond to a question. More implementation and training details are elaborated in Appendix~\ref{app:implement}.

\subsection{Datasets and Evaluation Metrics}

We experiment on two publicly available benchmark datasets commonly used for personalized dialogue evaluation: \texttt{MSC}~\citep{xu-etal-2022-beyond} and \texttt{LongMemEval}~\citep{wu2024longmemeval}. Additional details about datasets can be found in Appendix~\ref{app:dataset}.

For \texttt{MSC}, the evaluation measures if the generated response matches the human-provided ground truth. We follow \citet{li2024hello} to use \textbf{METEOR}~\citep{banerjee2005meteor} for measuring lexical similarity and \textbf{BERTScore}~\citep{Zhang2020BERTScore} for measuring semantic similarity. We also provide LLM judge results in Appendix~\ref{app:msc_llm}.

For \texttt{LongMemEval}, we follow the original paper to use \textbf{Recall@}$\mathbf{K}$ to evaluate the model's ability to retrieve relevant information for the query from conversation histories and use an LLM judge to measure the \textbf{Accuracy} of the generated answer by comparing it to the human-provided ground truth using Gemini-1.5-Pro. The prompt is presented in Appendix~\ref{app:judge}.

\subsection{Compared Methods}
To benchmark the performance of RMM, we compare it against the following baselines which represent different strategies for managing and retrieving long-term conversational memories, allowing for a comprehensive comparison with RMM.
\begin{itemize}[leftmargin=*,itemsep=1pt]
    \item \textbf{No History}: No history session is used.
    \item \textbf{Long Context}: This method directly incorporate as much conversation history as possible into the context window. Older turns are truncated.
    \item \textbf{RAG}: These models retrieve relevant turns or sessions for a given user query, concatenate them with the query, and feed the resulting input to the LLM for response generation. We use turns as the default granularity for better performance.
    \item \textbf{Personalized Dialogue Agents}: We consider two agent systems: (1) {MemoryBank}~\citep{zhong2024memorybank} treats conversation history as a fixed database and modulates retrieval using heuristics based on the forgetting curve.
    (2) {LD-Agent}~\citep{li2024hello} employs fixed conversation databases with additional retrieval modulation using strategies such as keywords matching.
\end{itemize}

\section{Experimental Results}

\begin{table*}[t]
    \begin{center}
    \scalebox{0.8}{
    \begin{tabular}{@{}lccccc@{}}
    \toprule
    \multirow{2}{*}{\textbf{Method}}      & \multirow{2}{*}{\textbf{Retriever}} & \multicolumn{2}{c}{\texttt{MSC}}       & \multicolumn{2}{c}{\texttt{LongMemEval}} \\ \cmidrule(l){3-6} 
                                         &                                     & \textbf{METEOR $(\%) \uparrow$} & \textbf{BERT $(\%) \uparrow$} & \textbf{Recall@5 $(\%) \uparrow$} & \textbf{Acc. $(\%) \uparrow$} \\ \midrule
No History & -                                       & 5.2         & 10.6      & -              & 0.0          \\
Long Context    & -                                  & 14.8        & 31.9      & -              & 57.4         \\ \midrule
    \multirow{3}{*}{RAG}                 & Contriever                          & 24.8                      & 50.8                    & 54.3                        & 58.8                    \\
                                         & Stella                              & 26.2                      & 51.6                    & 59.2                        & 61.4                    \\
                                         & GTE                                 & 27.5                      & 52.1                    & 62.4                        & 63.6                    \\ \midrule
    MemoryBank                           & Specific1                           & 20.1                      & 40.3                    & 58.6                           & 59.6                       \\
    LD-Agent                             & Specific2                           & 25.4                      & 51.5                    & 56.8                           & 59.2                       \\ \midrule
    \multirow{3}{*}{\textbf{RMM (Ours)}} & Contriever                          & 30.8                      & 55.4                    & 60.4                        & 61.2                    \\
                                         & Stella                              & 31.9                      & 56.3                    & 65.9                        & 64.8                    \\
                                         & GTE                                 & \textbf{33.4}                      & \textbf{57.1}                    & \textbf{69.8}                        & \textbf{70.4}                    \\ \midrule\midrule
    RAG                                  & Oracle                              & -                         & -                       & 100.0                         & 90.2                    \\ 
    \bottomrule
    \end{tabular}
    }
    \caption{Performance comparison of RMM with baseline methods on the \texttt{MSC} and \texttt{LongMemEval} datasets. Metrics include METEOR and BERT Scores for \texttt{MSC}, and Recall@5 and Accuracy (Acc.) scores for \texttt{LongMemEval}. RMM demonstrates superior performance across all metrics, highlighting its effectiveness in retrieval relevance and personalized response generation. No oracle retrieval is available for the \texttt{MSC} dataset. MemoryBank and LD-Agent utilize their specific methods for retrieval. 
    Scores are averaged over 3 runs and are reported in percentage ($\%$).}\label{tab:compare}
    \end{center}
    \end{table*}

\subsection{Main Results}
We present the main results shown in Table~\ref{tab:compare} and analyze each method's performance as follow.

\noindent\textbf{History matters:}~Without any history, the LLM performs poorly, achieving a METEOR score of 5.2$\%$ on \texttt{MSC} and 0.0$\%$ accuracy on \texttt{LongMemEval}, showing the necessity of historical context. 

\noindent\textbf{Long context is not enough:}  
Long-context models struggle due to fixed context windows and the inclusion of noisy context. On \texttt{MSC}, scores remain low (\textit{e.g.}, METEOR below 20$\%$, BERT score under 40$\%$), and on \texttt{LongMemEval}, accuracy is lower than 58$\%$. This limitation highlights their inability to retain and utilize long-term knowledge. 

\noindent\textbf{RAG Models:}  
RAG models outperform Long-Context LLMs by only incorporating relevant histories. With strong retrievers like GTE, RAG achieves 27.5$\%$ METEOR and 52.1$\%$ BERT Scores on \texttt{MSC} and 62.4$\%$ recall and 63.6$\%$ accuracy on \texttt{LongMemEval}. We also observe that the performance is retriever-dependent, where stronger retrievers boost the performance.

\noindent\textbf{Personalized Dialogue Agents:}  
MemoryBank and LD-Agent show more moderate improvements over Long-Context LLMs. For instance, LD-Agent achieves 25.4$\%$ METEOR and 51.5$\%$ BERT score on \texttt{MSC}, but these models fall short of RAG and RMM. Their reliance on heuristic-based retrieval potentially limits adaptability to complex tasks.

\noindent\textbf{Proposed RMM Framework:}  
RMM consistently achieves the best results across datasets and metrics. With GTE, RMM achieves 33.4$\%$ METEOR and 57.1$\%$ BERT on \texttt{MSC}, and 69.8$\%$ recall and 70.4$\%$ accuracy on \texttt{LongMemEval}. Even with weaker retrievers like Contriever, RMM maintains competitive performance, demonstrating robustness. The improvements stem from
RMM’s ability to integrate dynamic memory management with adaptive retrieval optimization enables it to retrieve and utilize relevant knowledge effectively, outperforming all baselines.

To further assess the impact of memory integration, we calculate the proportion of test examples where memory improves response quality. On \texttt{MSC}, memory improves on 86\% of responses, as the dataset frequently requires recalling prior discussion topics. On \texttt{LongMemEval}, where questions are deliberately designed to test historical recall, memory contributes to quality improvements in 100\% of cases. These results show the necessity of memory mechanisms in maintaining long-term coherence.
We provide case studies of the memory usage in Appendix~\ref{app:case}. 

\subsection{Ablation Study}
We conduct ablation study to evaluate the contributions of key components in the RMM framework. we present the results in Table~\ref{tab:ablation} and list our observations as below.

\begin{wraptable}{r}{0.5\textwidth}
\resizebox{\linewidth}{!}{
\begin{tabular}{@{}lcccc@{}}
\toprule
\multicolumn{1}{c}{\multirow{2}{*}{\textbf{Variant}}} & \multicolumn{2}{c}{\texttt{MSC}} & \multicolumn{2}{c}{\texttt{LongMemEval}} \\ \cmidrule(l){2-5} 
\multicolumn{1}{c}{}                                   & \textbf{METEOR}       & \textbf{BERT}        & \textbf{Recall@5}           & \textbf{Acc.}            \\ 
\midrule
    RAG                                        & 24.8        & 50.8      & 54.3             & 58.8         \\ \midrule
+ PR                                   & 28.6        & 53.3      & 57.4             & 59.6         \\
+ RR (W/O reranker)                                   & 20.3        & 31.8      & 34.2             & 31.0         \\
+ RR                       & 27.5        & 52.2      & 58.8             & 60.2         \\ \midrule
\textbf{RMM}            & \textbf{30.8}        & \textbf{55.4}      & \textbf{60.4}            & \textbf{61.2}         \\ \bottomrule
\end{tabular}}
\caption{Ablation study on the datasets. Variants evaluate the impact of key components in RMM: Prospective Reflection (PR), Retrospective Reflection (RR), and the reranker. RR (W/O reranker) means the retriever is fine-tuned instead.
Scores are obtained with Contriever and Gemini-1.5-Flash and in percentage ($\%$).}\label{tab:ablation}
\end{wraptable}

(\textit{\textbf{i}})~Adding Prospective Reflection boosts performance by organizing the memory into structured topics, which reduces redundancy and improves relevance. (\textit{\textbf{ii}})~Retrospective Reflection alone without a reranker misaligns retrieved content, leading to suboptimal results. Directly updating the retriever using RL rewards requires extensive amounts of training data for effective full fine-tuning, which is often difficult to obtain in real-world scenarios. Without sufficient data, it can lead to issues like catastrophic forgetting \citep{mccloskey1989catastrophic}. (\textit{\textbf{iii}})~The addition of the reranker alongside RR significantly enhances alignment, achieving 27.5$\%$ METEOR and 58.8$\%$ Recall@5, demonstrating its effectiveness in refining retrieval quality. (\textit{\textbf{iv}})~Finally, the complete RMM framework, which integrates Prospective Reflection, Retrospective Reflection, and the reranker, achieves the best results across all metrics, with a METEOR score of 30.8$\%$ on \texttt{MSC} and 60.4$\%$ Recall@5 on \texttt{LongMemEval}. This confirms that RMM enables more accurate and efficient future retrieval.

\subsection{Validation of Citation Scores}\label{sec:cite}

Our framework leverages LLM-generated citations to determine reward scores, guiding the retrieval refinement process. 
To assess the validity of the citation scores, we conduct evaluation on the \texttt{LongMemEval} dataset, using the \texttt{Gemini-1.5-Pro} model as the judge. The experiment tasks the LLM with determining whether cited memories were useful for response generation. The results, presented in Table~\ref{tab:citation_results}, demonstrate high precision, recall, and F1, confirming the effectiveness of citation-based scoring in our framework.

\begin{table}[b]
\begin{minipage}{.48\linewidth}
\centering

\resizebox{0.8\linewidth}{!}{
\begin{tabular}{@{}lccc@{}}
\toprule
\textbf{Metric}              & \textbf{Precision} & \textbf{Recall} & \textbf{F1} \\ \midrule
Useful memory     & 89.4               & 91.1            & 90.2        \\
Not useful memory & 87.2               & 84.6            & 85.9        \\ \midrule
Overall                      & 87.6               & 85.8            & 86.7        \\ \bottomrule
\end{tabular}}
\caption{Evaluation of citation-based scoring in RR for useful memory identification on \texttt{LongMemEval} (results in \%).}\label{tab:citation_results}

\end{minipage}
\hfill
\begin{minipage}{.48\linewidth}
\centering

\resizebox{\linewidth}{!}{
\begin{tabular}{ccccc}
\toprule
\multirow{2}{*}{\textbf{Method}} & \multirow{2}{*}{\textbf{LLM}} & \multicolumn{2}{c}{\texttt{MSC}} & \texttt{LongMemEval} \\
\cline{3-5} & & \textbf{METEOR} & \textbf{BERT} & \textbf{Acc.} \\
\midrule
\multirow{2}{*}{\makecell{Long\\Context}} & Gemini-1.5-Flash & 14.8 & 31.9 & \textbf{57.4} \\
& Gemini-1.5-Pro & \textbf{17.4} & \textbf{36.1} & 56.6 \\
\midrule
\multirow{2}{*}{\textbf{RMM}} & Gemini-1.5-Flash & \textbf{30.8} & \textbf{55.4} & \textbf{61.2} \\
& Gemini-1.5-Pro & 24.6 & 50.6 & 58.6 \\
\bottomrule
\end{tabular}
}
\caption{Effect of different LLMs on \texttt{MSC} and \texttt{LongMemEval}. Results (in \%) compare Long-Context LLMs and RMM using the Contriever retriever with Gemini models as generators.}\label{tab:llms}
\end{minipage}
\end{table}

\subsection{Effect of Different LLMs}
\label{sec:llm}

To examine the effect of different LLMs as generators, we evaluate both Gemini-1.5-Flash and Gemini-1.5-Pro in Long-Context LLMs and RMM. As shown in Table~\ref{tab:llms}, for Long-Context models, Gemini-1.5-Pro achieves slightly better performance than Gemini-1.5-Flash across all metrics, suggesting that a stronger model improves response quality when relying solely on extended context windows. However, for RMM, Gemini-1.5-Flash outperforms Gemini-1.5-Pro, achieving higher METEOR and BERT scores on \texttt{MSC} and better accuracy on \texttt{LongMemEval}. 
Similar observations are reported by \citet{wu2024longmemeval}, where GPT-4o-mini performs better than GPT-4o in personal knowledge QA. This trend can be attributed to stronger LLMs, such as Gemini-1.5-Pro, being more likely to abstain from answering queries involving personal information, possibly due to stronger alignment tuning aimed at enhancing privacy protection. 

\begin{figure}[t]
\begin{minipage}{.48\linewidth}
\centering
\includegraphics[width=1\linewidth]{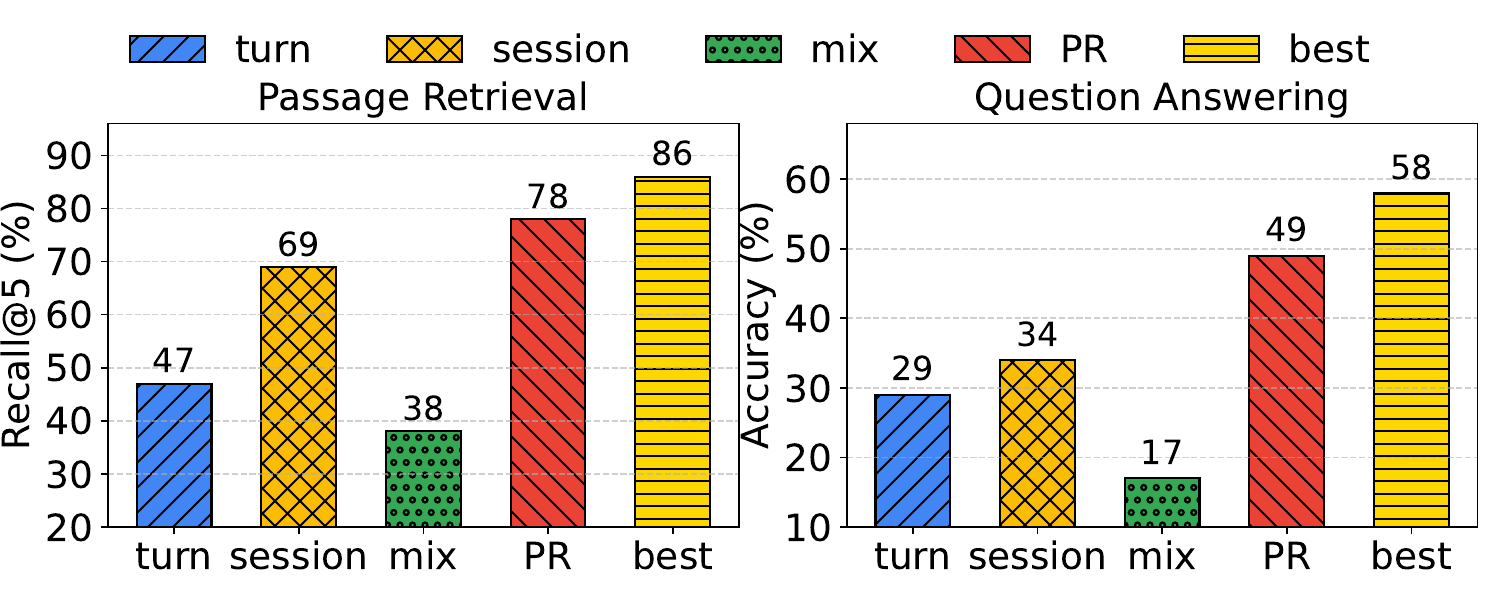}
\caption{Granularity analysis on randomly sampled 100 instances from \texttt{LongMemEval} with the GTE retriever and Gemini-1.5-Flash generator. ``\textcolor{bluecolor}{Turn}'' and ``\textcolor{orgcolor}{Session}'' indicate retrieval at a fixed granularity. ``\textcolor{ggreen}{Mix}'' represents retrieving from a pool combining both turns and sessions. ``\textcolor{red}{PR}'' refers to the granularity resulting from the proposed Prospective Reflection, while ``\textcolor{gyellow}{Best}'' corresponds to selecting the optimal granularity (either turn or session) for each instance.}
\label{fig:granu}
\end{minipage}
\hfill
\begin{minipage}{.48\linewidth}
\centering
\includegraphics[width=0.98\linewidth]{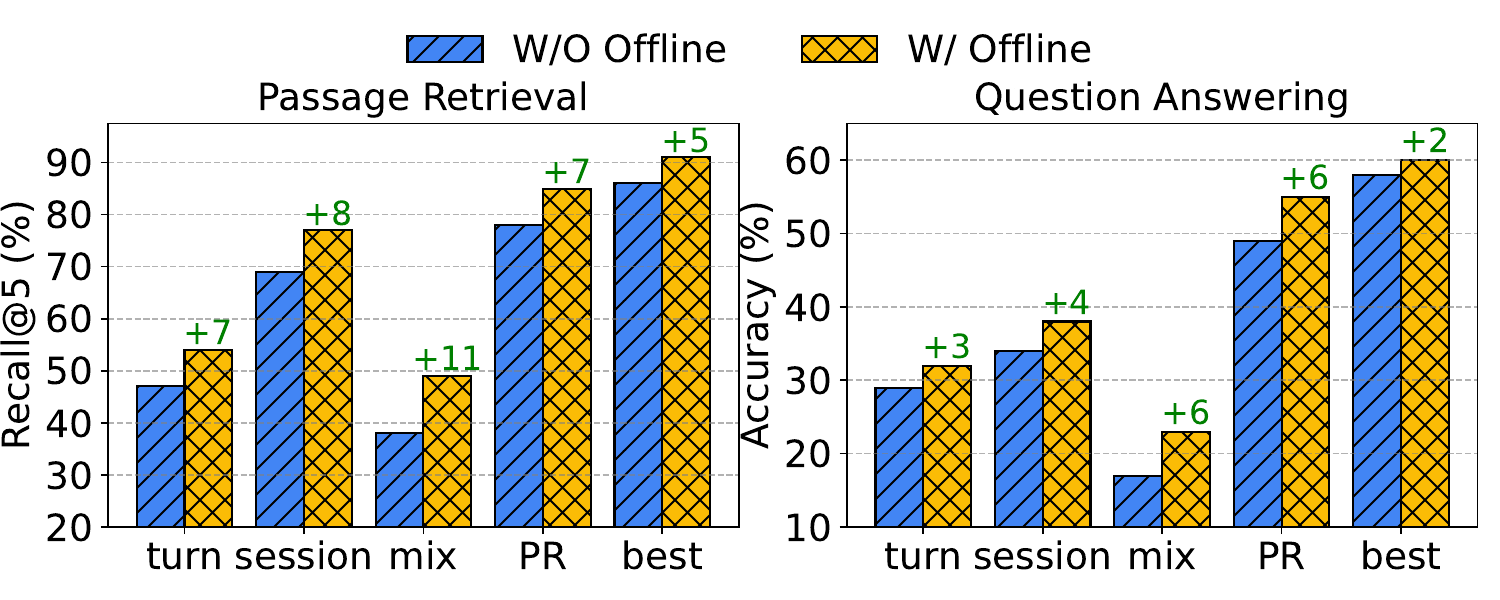}
\caption{Impact of offline pretraining on retriever performance for \texttt{LongMemEval} dataset with the same 100 random samples as Figure~\ref{fig:granu}. Results without offline pretraining are shown in \textcolor{bluecolor}{blue}, while results with offline pretraining are shown in \textcolor{orgcolor}{orange}. Offline pretraining improves recall and accuracy across all settings.}
\label{fig:offline}
\end{minipage}
\end{figure}

\subsection{Effect of Different Granularities}\label{sec:granu}

We conduct experiments to show the advantage of the flexible granularity resulting from the proposed Prospective Reflection (PR) over pre-defined fixed granularities as baselines. 
Results in Figure~\ref{fig:granu} show that fixed granularities, such as ``\textcolor{bluecolor}{turn}'' and ``\textcolor{orgcolor}{session}'', achieve moderate performance, with session-level retrieval outperforming turn-level due to richer contexts. The ``\textcolor{ggreen}{mixed}'' granularity underperforms, likely due to increased noise from a larger search space. The best configuration, which selects the optimal granularity per instance, achieves the highest scores, demonstrating the importance of adaptive memory organization. In contrast, \textcolor{red}{PR} improves performance by integrating fragmented conversational segments  into cohesive memory structure, exhibiting an approaching performance with the \textcolor{gyellow}{best} oracle granularity.

\subsection{Offline Supervised Training}\label{sec:offline}

We further investigate the applicability of RMM in scenarios where a handful of labelled retrieval data is available, allowing for offline supervised pretraining (based on the off-the-shelf retriever) before online refinement. Figure~\ref{fig:offline} illustrates the impact of offline pretraining on retriever performance on the \texttt{LongMemEval} dataset. We randomly select 100 samples as test data with the rest as training and validation sets and apply vanilla supervised contrastive learning for the GTE retriever~\citep{li2023towards}.
As the results show, across all settings, RMM consistently benefits from offline pretraining (\textcolor{orgcolor}{orange} bars) by outperforming retrievers without pretraining (\textcolor{bluecolor}{blue} bars). 
These results demonstrate that offline pretraining can enhance the retriever's ability to identify relevant information, providing a robust foundation for subsequent fine-tuning via RL.

\subsection{The Impact of Top-\textit{K} and Top-\textit{M} for RMM}\label{app:topk}

\begin{wraptable}{r}{0.5\textwidth}
\centering
\resizebox{\linewidth}{!}{
\begin{tabular}{@{}lccc|cc@{}}
\toprule
\multirow{2}{*}{\textbf{Model}}      & \multirow{2}{*}{\textbf{Retriever}} & \multicolumn{4}{c}{\texttt{LongMemEval}}                                                          \\ \cmidrule(l){3-6} 
                                     &                                     & \textbf{Recall@5} & \textbf{Acc.} & \textbf{Recall@10} & \textbf{Acc.} \\ \midrule
\multirow{3}{*}{RMM} & Contriever                          & 60.4                        & 61.2                    & 67.2                         & 66.8                    \\
                                     & Stella                              & 65.9                        & 64.8                    & 70.6                         & 71.0                    \\
                                     & GTE                                 & \textbf{69.8}               & \textbf{70.4}           & \textbf{74.4}                & \textbf{73.8}           \\ \bottomrule
\end{tabular}}
\caption{Impact of Top-$K$ (retrieved memories) and Top-$M$ (reranked memories) on \texttt{LongMemEval} performance. Results include Recall@5 (Top-$K$ = 20, Top-$M$ = 5) and Recall@10 (Top-$K$ = 50, Top-$M$ = 10), and corresponding Accuracy across different retrievers. Results show that increasing the number of retrieved and reranked memories improves retrieval and QA performance on the \texttt{LongMemEval} dataset.}\label{tab:topk}
\vspace{5mm}
\centering
\resizebox{\linewidth}{!}{%
\begin{tabular}{@{}ccccc@{}}
\toprule
\textbf{Method}      & \textbf{LLM}           & \textbf{METEOR} & \textbf{BERT} & \makecell{\textbf{LLM-as-a-Judge}\\ \textbf{(Yes\%)}} \\
\midrule
\multirow{2}{*}{\makecell{Long\\Context}} & Gemini-1.5-Flash & 14.8 & 31.9 & \textbf{25.4} \\
& Gemini-1.5-Pro & \textbf{17.4} & \textbf{36.1} & 22.8 \\
\midrule
\multirow{2}{*}{RMM} & Gemini-1.5-Flash & \textbf{30.8} & \textbf{55.4} & \textbf{69.7} \\
& Gemini-1.5-Pro & 24.6 & 50.6 & 65.4 \\
\bottomrule
\end{tabular}
}
\caption{Results for MSC with LLM-as-a-judge. RMM shows consistent advantages with more fine-grained evaluation.}
\vspace{5mm}
\centering
\resizebox{\linewidth}{!}{%
\begin{tabular}{lcc}
\toprule
\textbf{Pair} & \textbf{Cohen's Kappa ($\kappa$)} & \textbf{Interpretation} \\
\midrule
Human A vs. Human B & 0.82 & Substantial agreement \\
LLM vs. Human A     & 0.71 & Substantial agreement \\
LLM vs. Human B     & 0.69 & Substantial agreement \\
\bottomrule
\end{tabular}}
\caption{Cohen's Kappa Agreement Between Annotators and LLM}\label{tab:huamn}
\vspace{-2cm}
\end{wraptable}

The results in Table~\ref{tab:topk} evaluate the impact of the number of retrieved memories (Top-$K$) and the number of reranked memories used for response generation (Top-$M$) in the RMM framework. Specifically, we analyze the performance on \texttt{LongMemEval} using Recall@5 (Top-$K$ = 20, Top-$M$ = 5), Recall@10 (Top-$K$ = 50, Top-$M$ = 10), and their corresponding QA accuracy scores.

The results demonstrate two key findings.
First, increasing the number of memories (\textit{M}) from 5 to 10 consistently improves both retrieval and accuracy metrics across all retrievers. For example, with the GTE retriever, Recall improves from 69.8\% to 74.4\%, and Accuracy increases from 70.4\% to 73.8\%.  
Second, the performance gain is most significant for stronger retrievers like GTE and Stella, highlighting the importance of retrieval quality. RMM with GTE achieves the best results of 70.4\% Accuracy with Top-$K$ = 20, Top-$M$ = 5 and 73.8\% Accuracy with Top-$K$ = 50, Top-$M$ = 10.

These observations emphasize that careful selection of Top-$K$ and Top-$M$ values can enhance both retrieval relevance and downstream QA performance. The combination of effective retrieval and reranking ensures that RMM efficiently leverages the most relevant information for long-term dialogue tasks.

\subsection{Results for \texttt{MSC} with LLM-as-a-judge}
\label{app:msc_llm}

For fair comparison, we follow prior work~\citep{xu-etal-2022-beyond,wu2024longmemeval} and use METEOR and BERTScore. Here we include additional results using LLM-as-a-judge. Following \citet{wu2024longmemeval}, we use Gemini-1.5-Pro to decide whether the generated answer matches the ground-truth as a binary annotation. The prompt we used in given in Appendix~\ref{app:judge}. LLM-as-a-judge results also show the effectiveness of the proposed RMM. 

Furthermore, we conducted a human evaluation on 100 randomly sampled instances from the MSC dataset. 
For each instance, annotators were shown the user query, the ground-truth response, and the model-generated response—mirroring the setup used for the LLM-based evaluations. We adopted the same instruction used in our LLM judge prompt, and asked human annotators to answer Yes/No to evaluate each instance. The annotation was independently performed by two NLP researchers. Both are PhD students working in NLP and not on the author list. We report inter-annotator agreement and agreement between human judgments and the LLM judge in Table~\ref{tab:huamn}.


\section{Conclusion}

We present {RMM}, a framework that integrates {Prospective Reflection} for structured, topic-based memory organization and {Retrospective Reflection} for dynamic memory reranking via reinforcement learning. Experimental results on benchmark datasets demonstrate that RMM outperforms state-of-the-art baselines in retrieval relevance and response quality for personalized dialogue tasks. 
By identifying limitations in existing memory management approaches—particularly those relying on fixed granularity and static retrievers, we highlight key challenges and avenues for future research in long-term dialogue memory modeling.

\section*{Limitations}
While the proposed RMM framework demonstrates significant improvements in retrieval relevance and response quality, it is not without limitations. First, RMM relies on reinforcement learning for memory reranking, which can be computationally expensive, especially for large-scale datasets or real-time applications. Second, the current framework primarily focuses on textual data, limiting its applicability to multi-modal dialogue systems that incorporate images, audio, or video. Additionally, the memory updating mechanism may require further optimization to handle dynamically evolving long-term user interactions efficiently.

For future work, we plan to address these limitations by exploring more efficient reinforcement learning techniques and lightweight memory reranking strategies. We also aim to extend RMM to multi-modal dialogue systems to accommodate diverse user interactions. Furthermore, we will investigate privacy-preserving techniques to ensure safe deployment of RMM in real-world personalized dialogue applications where sensitive user data is involved.

\section*{Ethical Statement}
This work focuses on developing a framework for long-term personalized dialogue systems to improve user experiences. However, we acknowledge the potential ethical implications of handling personal data in such systems. The RMM framework relies on historical conversations, which may contain sensitive or private information. To mitigate privacy risks, we recommend adopting robust encryption and privacy-preserving methods, such as differential privacy or federated learning, during data collection and model training.

Additionally, we emphasize the importance of transparent data usage policies and obtaining user consent when deploying personalized dialogue systems. Efforts should also be made to minimize biases in memory retrieval and response generation to ensure fairness and inclusivity across diverse user groups. Future work will continue to prioritize ethical considerations to promote the responsible development and deployment of personalized dialogue technologies.

\bibliographystyle{abbrvnat}
\nobibliography*
\bibliography{custom}

\begin{thebibliography}{53}
\providecommand{\natexlab}[1]{#1}
\providecommand{\url}[1]{\texttt{#1}}
\expandafter\ifx\csname urlstyle\endcsname\relax
  \providecommand{\doi}[1]{doi: #1}\else
  \providecommand{\doi}{doi: \begingroup \urlstyle{rm}\Url}\fi

\bibitem[Bae et~al.(2022)Bae, Kwak, Kang, Lee, Kim, Jeong, Kim, Lee, Park, and
  Sung]{bae2022keep}
S.~Bae, D.~Kwak, S.~Kang, M.~Y. Lee, S.~Kim, Y.~Jeong, H.~Kim, S.-W. Lee,
  W.~Park, and N.~Sung.
\newblock Keep me updated! memory management in long-term conversations.
\newblock In Y.~Goldberg, Z.~Kozareva, and Y.~Zhang, editors, \emph{Findings of
  the Association for Computational Linguistics: EMNLP 2022}, pages 3769--3787,
  Abu Dhabi, United Arab Emirates, 2022. Association for Computational
  Linguistics.
\newblock \doi{10.18653/v1/2022.findings-emnlp.276}.
\newblock URL \url{https://aclanthology.org/2022.findings-emnlp.276}.

\bibitem[Banerjee and Lavie(2005)]{banerjee2005meteor}
S.~Banerjee and A.~Lavie.
\newblock {METEOR}: An automatic metric for {MT} evaluation with improved
  correlation with human judgments.
\newblock In J.~Goldstein, A.~Lavie, C.-Y. Lin, and C.~Voss, editors,
  \emph{Proceedings of the {ACL} Workshop on Intrinsic and Extrinsic Evaluation
  Measures for Machine Translation and/or Summarization}, pages 65--72, Ann
  Arbor, Michigan, 2005. Association for Computational Linguistics.
\newblock URL \url{https://aclanthology.org/W05-0909}.

\bibitem[Buchmann et~al.(2024)Buchmann, Liu, and
  Gurevych]{buchmann-etal-2024-attribute}
J.~Buchmann, X.~Liu, and I.~Gurevych.
\newblock Attribute or abstain: Large language models as long document
  assistants.
\newblock In Y.~Al-Onaizan, M.~Bansal, and Y.-N. Chen, editors,
  \emph{Proceedings of the 2024 Conference on Empirical Methods in Natural
  Language Processing}, pages 8113--8140, Miami, Florida, USA, 2024.
  Association for Computational Linguistics.
\newblock \doi{10.18653/v1/2024.emnlp-main.463}.
\newblock URL \url{https://aclanthology.org/2024.emnlp-main.463}.

\bibitem[Chen et~al.(2024)Chen, Liu, Huang, Wu, Liu, Jiang, Pu, Lei, Chen,
  Wang, Zheng, Lian, and Chen]{chen2024large}
J.~Chen, Z.~Liu, X.~Huang, C.~Wu, Q.~Liu, G.~Jiang, Y.~Pu, Y.~Lei, X.~Chen,
  X.~Wang, K.~Zheng, D.~Lian, and E.~Chen.
\newblock When large language models meet personalization: perspectives of
  challenges and opportunities.
\newblock \emph{World Wide Web}, 27\penalty0 (4), June 2024.
\newblock ISSN 1386-145X.
\newblock \doi{10.1007/s11280-024-01276-1}.
\newblock URL \url{https://doi.org/10.1007/s11280-024-01276-1}.

\bibitem[Chhikara et~al.(2025)Chhikara, Khant, Aryan, Singh, and
  Yadav]{chhikara2025mem0buildingproductionreadyai}
P.~Chhikara, D.~Khant, S.~Aryan, T.~Singh, and D.~Yadav.
\newblock Mem0: Building production-ready ai agents with scalable long-term
  memory, 2025.
\newblock URL \url{https://arxiv.org/abs/2504.19413}.

\bibitem[Dong et~al.(2024)Dong, Hu, and Collier]{dong2024can}
Y.~R. Dong, T.~Hu, and N.~Collier.
\newblock Can {LLM} be a personalized judge?
\newblock In Y.~Al-Onaizan, M.~Bansal, and Y.-N. Chen, editors, \emph{Findings
  of the Association for Computational Linguistics: EMNLP 2024}, pages
  10126--10141, Miami, Florida, USA, Nov. 2024. Association for Computational
  Linguistics.
\newblock \doi{10.18653/v1/2024.findings-emnlp.592}.
\newblock URL \url{https://aclanthology.org/2024.findings-emnlp.592/}.

\bibitem[Guan et~al.(2024)Guan, Wang, Chu, Wang, Ni, Song, and
  Zhuang]{guan2023intelligent}
Y.~Guan, D.~Wang, Z.~Chu, S.~Wang, F.~Ni, R.~Song, and C.~Zhuang.
\newblock Intelligent agents with llm-based process automation.
\newblock In \emph{Proceedings of the 30th ACM SIGKDD Conference on Knowledge
  Discovery and Data Mining}, KDD '24, page 5018–5027, New York, NY, USA,
  2024. Association for Computing Machinery.
\newblock ISBN 9798400704901.
\newblock \doi{10.1145/3637528.3671646}.
\newblock URL \url{https://doi.org/10.1145/3637528.3671646}.

\bibitem[Gumbel(1954)]{gumbel1948statistical}
E.~Gumbel.
\newblock \emph{Statistical Theory of Extreme Values and Some Practical
  Applications: A Series of Lectures}.
\newblock Applied mathematics series. U.S. Government Printing Office, 1954.
\newblock URL \url{https://books.google.com/books?id=SNpJAAAAMAAJ}.

\bibitem[Izacard et~al.(2022)Izacard, Caron, Hosseini, Riedel, Bojanowski,
  Joulin, and Grave]{izacard2021unsupervised}
G.~Izacard, M.~Caron, L.~Hosseini, S.~Riedel, P.~Bojanowski, A.~Joulin, and
  E.~Grave.
\newblock Unsupervised dense information retrieval with contrastive learning.
\newblock \emph{Transactions on Machine Learning Research}, 2022.
\newblock ISSN 2835-8856.
\newblock URL \url{https://openreview.net/forum?id=jKN1pXi7b0}.

\bibitem[Jang et~al.(2017)Jang, Gu, and Poole]{jang2016categorical}
E.~Jang, S.~Gu, and B.~Poole.
\newblock Categorical reparameterization with gumbel-softmax.
\newblock In \emph{5th International Conference on Learning Representations,
  {ICLR} 2017, Toulon, France, April 24-26, 2017, Conference Track
  Proceedings}. OpenReview.net, 2017.
\newblock URL \url{https://openreview.net/forum?id=rkE3y85ee}.

\bibitem[Jiang et~al.(2025)Jiang, Sun, Liang, and Zhang]{jiang2024retrieve}
Z.~Jiang, M.~Sun, L.~Liang, and Z.~Zhang.
\newblock Retrieve, summarize, plan: Advancing multi-hop question answering
  with an iterative approach.
\newblock In \emph{Companion Proceedings of the ACM on Web Conference 2025},
  WWW '25, page 1677–1686, New York, NY, USA, 2025. Association for Computing
  Machinery.
\newblock ISBN 9798400713316.
\newblock \doi{10.1145/3701716.3716889}.
\newblock URL \url{https://doi.org/10.1145/3701716.3716889}.

\bibitem[Kenthapadi et~al.(2024)Kenthapadi, Sameki, and
  Taly]{kenthapadi2024grounding}
K.~Kenthapadi, M.~Sameki, and A.~Taly.
\newblock Grounding and evaluation for large language models: Practical
  challenges and lessons learned (survey).
\newblock In \emph{Proceedings of the 30th ACM SIGKDD Conference on Knowledge
  Discovery and Data Mining}, KDD '24, page 6523–6533, New York, NY, USA,
  2024. Association for Computing Machinery.
\newblock ISBN 9798400704901.
\newblock \doi{10.1145/3637528.3671467}.
\newblock URL \url{https://doi.org/10.1145/3637528.3671467}.

\bibitem[Kolasani(2023)]{kolasani2023optimizing}
S.~Kolasani.
\newblock Optimizing natural language processing, large language models (llms)
  for efficient customer service, and hyper-personalization to enable
  sustainable growth and revenue.
\newblock \emph{Transactions on Latest Trends in Artificial Intelligence},
  4\penalty0 (4), 2023.
\newblock ISSN 3246-548X.
\newblock URL \url{https://ijsdcs.com/index.php/TLAI/article/view/476}.

\bibitem[Lee et~al.(2023)Lee, Hartmann, Park, Papailiopoulos, and
  Lee]{lee2023prompted}
G.~Lee, V.~Hartmann, J.~Park, D.~Papailiopoulos, and K.~Lee.
\newblock Prompted {LLM}s as chatbot modules for long open-domain conversation.
\newblock In A.~Rogers, J.~Boyd-Graber, and N.~Okazaki, editors, \emph{Findings
  of the Association for Computational Linguistics: ACL 2023}, pages
  4536--4554, Toronto, Canada, 2023. Association for Computational Linguistics.
\newblock \doi{10.18653/v1/2023.findings-acl.277}.
\newblock URL \url{https://aclanthology.org/2023.findings-acl.277}.

\bibitem[Li et~al.(2024{\natexlab{a}})Li, Verga, Sen, Yang, Viswanathan, Lewis,
  Watanabe, and Su]{li2024alr}
H.~Li, P.~Verga, P.~Sen, B.~Yang, V.~Viswanathan, P.~Lewis, T.~Watanabe, and
  Y.~Su.
\newblock Alr$^2$: A retrieve-then-reason framework for long-context question
  answering.
\newblock \emph{ArXiv preprint}, abs/2410.03227, 2024{\natexlab{a}}.
\newblock URL \url{https://arxiv.org/abs/2410.03227}.

\bibitem[Li et~al.(2025)Li, Yang, Zhang, Deng, Wang, and Chua]{li2024hello}
H.~Li, C.~Yang, A.~Zhang, Y.~Deng, X.~Wang, and T.-S. Chua.
\newblock Hello again! {LLM}-powered personalized agent for long-term dialogue.
\newblock In L.~Chiruzzo, A.~Ritter, and L.~Wang, editors, \emph{Proceedings of
  the 2025 Conference of the Nations of the Americas Chapter of the Association
  for Computational Linguistics: Human Language Technologies (Volume 1: Long
  Papers)}, pages 5259--5276, Albuquerque, New Mexico, Apr. 2025. Association
  for Computational Linguistics.
\newblock ISBN 979-8-89176-189-6.
\newblock \doi{10.18653/v1/2025.naacl-long.272}.
\newblock URL \url{https://aclanthology.org/2025.naacl-long.272/}.

\bibitem[Li et~al.(2024{\natexlab{b}})Li, Wen, Wang, Li, Yuan, Liu, Liu, Xu,
  Wang, Sun, et~al.]{li2024personal}
Y.~Li, H.~Wen, W.~Wang, X.~Li, Y.~Yuan, G.~Liu, J.~Liu, W.~Xu, X.~Wang, Y.~Sun,
  et~al.
\newblock Personal llm agents: Insights and survey about the capability,
  efficiency and security.
\newblock \emph{ArXiv preprint}, abs/2401.05459, 2024{\natexlab{b}}.
\newblock URL \url{https://arxiv.org/abs/2401.05459}.

\bibitem[LI et~al.(2025)LI, Jiang, Wu, Luo, Ahn, Zhang, Abdi, Li, Gao, Yang,
  and Qiu]{li2024scbench}
Y.~LI, H.~Jiang, Q.~Wu, X.~Luo, S.~Ahn, C.~Zhang, A.~H. Abdi, D.~Li, J.~Gao,
  Y.~Yang, and L.~Qiu.
\newblock {SCB}ench: A {KV} cache-centric analysis of long-context methods.
\newblock In \emph{The Thirteenth International Conference on Learning
  Representations}, 2025.
\newblock URL \url{https://openreview.net/forum?id=gkUyYcY1W9}.

\bibitem[Li et~al.(2023)Li, Zhang, Zhang, Long, Xie, and Zhang]{li2023towards}
Z.~Li, X.~Zhang, Y.~Zhang, D.~Long, P.~Xie, and M.~Zhang.
\newblock Towards general text embeddings with multi-stage contrastive
  learning.
\newblock \emph{ArXiv preprint}, abs/2308.03281, 2023.
\newblock URL \url{https://arxiv.org/abs/2308.03281}.

\bibitem[Liu et~al.(2024{\natexlab{a}})Liu, Zaharia, and Abbeel]{liu2023ring}
H.~Liu, M.~Zaharia, and P.~Abbeel.
\newblock Ringattention with blockwise transformers for near-infinite context.
\newblock In \emph{The Twelfth International Conference on Learning
  Representations}, 2024{\natexlab{a}}.
\newblock URL \url{https://openreview.net/forum?id=WsRHpHH4s0}.

\bibitem[Liu et~al.(2024{\natexlab{b}})Liu, Lin, Hewitt, Paranjape, Bevilacqua,
  Petroni, and Liang]{liu2024lost}
N.~F. Liu, K.~Lin, J.~Hewitt, A.~Paranjape, M.~Bevilacqua, F.~Petroni, and
  P.~Liang.
\newblock Lost in the middle: How language models use long contexts.
\newblock \emph{Transactions of the Association for Computational Linguistics},
  12:\penalty0 157--173, 2024{\natexlab{b}}.
\newblock \doi{10.1162/tacl_a_00638}.
\newblock URL \url{https://aclanthology.org/2024.tacl-1.9}.

\bibitem[Liu et~al.(2025)Liu, Tang, Dong, Li, Li, Hu, and Chu]{liu2025chunkkv}
X.~Liu, Z.~Tang, P.~Dong, Z.~Li, B.~Li, X.~Hu, and X.~Chu.
\newblock Chunkkv: Semantic-preserving kv cache compression for efficient
  long-context llm inference.
\newblock \emph{ArXiv preprint}, abs/2502.00299, 2025.
\newblock URL \url{https://arxiv.org/abs/2502.00299}.

\bibitem[Lu et~al.(2023)Lu, An, Lin, Pergola, He, Yin, Sun, and
  Wu]{lu2023memochat}
J.~Lu, S.~An, M.~Lin, G.~Pergola, Y.~He, D.~Yin, X.~Sun, and Y.~Wu.
\newblock Memochat: Tuning llms to use memos for consistent long-range
  open-domain conversation.
\newblock \emph{ArXiv preprint}, abs/2308.08239, 2023.
\newblock URL \url{https://arxiv.org/abs/2308.08239}.

\bibitem[Maddison et~al.(2014)Maddison, Tarlow, and
  Minka]{maddison2014sampling}
C.~J. Maddison, D.~Tarlow, and T.~Minka.
\newblock A* sampling.
\newblock In Z.~Ghahramani, M.~Welling, C.~Cortes, N.~Lawrence, and
  K.~Weinberger, editors, \emph{Advances in Neural Information Processing
  Systems}, volume~27. Curran Associates, Inc., 2014.
\newblock URL
  \url{https://proceedings.neurips.cc/paper_files/paper/2014/file/309fee4e541e51de2e41f21bebb342aa-Paper.pdf}.

\bibitem[Maharana et~al.(2024)Maharana, Lee, Tulyakov, Bansal, Barbieri, and
  Fang]{maharana2024evaluating}
A.~Maharana, D.-H. Lee, S.~Tulyakov, M.~Bansal, F.~Barbieri, and Y.~Fang.
\newblock Evaluating very long-term conversational memory of {LLM} agents.
\newblock In L.-W. Ku, A.~Martins, and V.~Srikumar, editors, \emph{Proceedings
  of the 62nd Annual Meeting of the Association for Computational Linguistics
  (Volume 1: Long Papers)}, pages 13851--13870, Bangkok, Thailand, Aug. 2024.
  Association for Computational Linguistics.
\newblock \doi{10.18653/v1/2024.acl-long.747}.
\newblock URL \url{https://aclanthology.org/2024.acl-long.747/}.

\bibitem[McCloskey and Cohen(1989)]{mccloskey1989catastrophic}
M.~McCloskey and N.~J. Cohen.
\newblock Catastrophic interference in connectionist networks: The sequential
  learning problem.
\newblock volume~24 of \emph{Psychology of Learning and Motivation}, pages
  109--165. Academic Press, 1989.
\newblock \doi{https://doi.org/10.1016/S0079-7421(08)60536-8}.
\newblock URL
  \url{https://www.sciencedirect.com/science/article/pii/S0079742108605368}.

\bibitem[Mendon{\c{c}}a et~al.(2024)Mendon{\c{c}}a, Lavie, and
  Trancoso]{mendoncca2024benchmarking}
J.~Mendon{\c{c}}a, A.~Lavie, and I.~Trancoso.
\newblock On the benchmarking of {LLM}s for open-domain dialogue evaluation.
\newblock In E.~Nouri, A.~Rastogi, G.~Spithourakis, B.~Liu, Y.-N. Chen, Y.~Li,
  A.~Albalak, H.~Wakaki, and A.~Papangelis, editors, \emph{Proceedings of the
  6th Workshop on NLP for Conversational AI (NLP4ConvAI 2024)}, pages 1--12,
  Bangkok, Thailand, Aug. 2024. Association for Computational Linguistics.
\newblock URL \url{https://aclanthology.org/2024.nlp4convai-1.1/}.

\bibitem[Ong et~al.(2025)Ong, Kim, Gwak, Chae, Kwon, Jo, Hwang, Lee, and
  Yeo]{kim2024theanine}
K.~T.-i. Ong, N.~Kim, M.~Gwak, H.~Chae, T.~Kwon, Y.~Jo, S.-w. Hwang, D.~Lee,
  and J.~Yeo.
\newblock Towards lifelong dialogue agents via timeline-based memory
  management.
\newblock In L.~Chiruzzo, A.~Ritter, and L.~Wang, editors, \emph{Proceedings of
  the 2025 Conference of the Nations of the Americas Chapter of the Association
  for Computational Linguistics: Human Language Technologies (Volume 1: Long
  Papers)}, pages 8631--8661, Albuquerque, New Mexico, Apr. 2025. Association
  for Computational Linguistics.
\newblock ISBN 979-8-89176-189-6.
\newblock \doi{10.18653/v1/2025.naacl-long.435}.
\newblock URL \url{https://aclanthology.org/2025.naacl-long.435/}.

\bibitem[Packer et~al.(2024)Packer, Wooders, Lin, Fang, Patil, Stoica, and
  Gonzalez]{packer2024memgptllmsoperatingsystems}
C.~Packer, S.~Wooders, K.~Lin, V.~Fang, S.~G. Patil, I.~Stoica, and J.~E.
  Gonzalez.
\newblock Memgpt: Towards llms as operating systems, 2024.
\newblock URL \url{https://arxiv.org/abs/2310.08560}.

\bibitem[Pan et~al.(2025)Pan, Wu, Jiang, Luo, Cheng, Li, Yang, Lin, Zhao, Qiu,
  and Gao]{pansecom}
Z.~Pan, Q.~Wu, H.~Jiang, X.~Luo, H.~Cheng, D.~Li, Y.~Yang, C.-Y. Lin, H.~V.
  Zhao, L.~Qiu, and J.~Gao.
\newblock Secom: On memory construction and retrieval for personalized
  conversational agents.
\newblock In \emph{The Thirteenth International Conference on Learning
  Representations}, 2025.
\newblock URL \url{https://openreview.net/forum?id=xKDZAW0He3}.

\bibitem[Pei et~al.(2021)Pei, Ren, and de~Rijke]{pei2021cooperative}
J.~Pei, P.~Ren, and M.~de~Rijke.
\newblock A cooperative memory network for personalized task-oriented dialogue
  systems with incomplete user profiles.
\newblock In J.~Leskovec, M.~Grobelnik, M.~Najork, J.~Tang, and L.~Zia,
  editors, \emph{{WWW} '21: The Web Conference 2021, Virtual Event / Ljubljana,
  Slovenia, April 19-23, 2021}, pages 1552--1561. {ACM} / {IW3C2}, 2021.
\newblock \doi{10.1145/3442381.3449843}.
\newblock URL \url{https://doi.org/10.1145/3442381.3449843}.

\bibitem[Rasmussen et~al.(2025)Rasmussen, Paliychuk, Beauvais, Ryan, and
  Chalef]{rasmussen2025zep}
P.~Rasmussen, P.~Paliychuk, T.~Beauvais, J.~Ryan, and D.~Chalef.
\newblock Zep: A temporal knowledge graph architecture for agent memory, 2025.
\newblock URL \url{https://arxiv.org/abs/2501.13956}.

\bibitem[Shi et~al.(2023)Shi, Chen, Misra, Scales, Dohan, Chi, Sch{\"{a}}rli,
  and Zhou]{shi2023large}
F.~Shi, X.~Chen, K.~Misra, N.~Scales, D.~Dohan, E.~H. Chi, N.~Sch{\"{a}}rli,
  and D.~Zhou.
\newblock Large language models can be easily distracted by irrelevant context.
\newblock In A.~Krause, E.~Brunskill, K.~Cho, B.~Engelhardt, S.~Sabato, and
  J.~Scarlett, editors, \emph{International Conference on Machine Learning,
  {ICML} 2023, 23-29 July 2023, Honolulu, Hawaii, {USA}}, volume 202 of
  \emph{Proceedings of Machine Learning Research}, pages 31210--31227. {PMLR},
  2023.
\newblock URL \url{https://proceedings.mlr.press/v202/shi23a.html}.

\bibitem[Tseng et~al.(2024)Tseng, Huang, Hsiao, Chen, Huang, Meng, and
  Chen]{tseng2024two}
Y.-M. Tseng, Y.-C. Huang, T.-Y. Hsiao, W.-L. Chen, C.-W. Huang, Y.~Meng, and
  Y.-N. Chen.
\newblock Two tales of persona in {LLM}s: A survey of role-playing and
  personalization.
\newblock In Y.~Al-Onaizan, M.~Bansal, and Y.-N. Chen, editors, \emph{Findings
  of the Association for Computational Linguistics: EMNLP 2024}, pages
  16612--16631, Miami, Florida, USA, Nov. 2024. Association for Computational
  Linguistics.
\newblock \doi{10.18653/v1/2024.findings-emnlp.969}.
\newblock URL \url{https://aclanthology.org/2024.findings-emnlp.969/}.

\bibitem[Walker et~al.(1997)Walker, Litman, Kamm, and
  Abella]{walker1997paradise}
M.~A. Walker, D.~J. Litman, C.~A. Kamm, and A.~Abella.
\newblock {PARADISE}: A framework for evaluating spoken dialogue agents.
\newblock In \emph{35th Annual Meeting of the Association for Computational
  Linguistics and 8th Conference of the {E}uropean Chapter of the Association
  for Computational Linguistics}, pages 271--280, Madrid, Spain, July 1997.
  Association for Computational Linguistics.
\newblock \doi{10.3115/976909.979652}.
\newblock URL \url{https://aclanthology.org/P97-1035/}.

\bibitem[Wang et~al.(2023)Wang, Ding, Cao, Tian, Wang, Tao, and
  Guo]{wang2023recursively}
Q.~Wang, L.~Ding, Y.~Cao, Z.~Tian, S.~Wang, D.~Tao, and L.~Guo.
\newblock Recursively summarizing enables long-term dialogue memory in large
  language models.
\newblock \emph{ArXiv preprint}, abs/2308.15022, 2023.
\newblock URL \url{https://arxiv.org/abs/2308.15022}.

\bibitem[Wang et~al.(2025)Wang, Mao, Fried, and
  Neubig]{wang2024agentworkflowmemory}
Z.~Z. Wang, J.~Mao, D.~Fried, and G.~Neubig.
\newblock Agent workflow memory.
\newblock In \emph{Forty-second International Conference on Machine Learning},
  2025.
\newblock URL \url{https://openreview.net/forum?id=NTAhi2JEEE}.

\bibitem[Weizenbaum(1966)]{weizenbaum1966eliza}
J.~Weizenbaum.
\newblock Eliza—a computer program for the study of natural language
  communication between man and machine.
\newblock \emph{Commun. ACM}, 9\penalty0 (1):\penalty0 36–45, Jan. 1966.
\newblock ISSN 0001-0782.
\newblock \doi{10.1145/365153.365168}.
\newblock URL \url{https://doi.org/10.1145/365153.365168}.

\bibitem[Wen et~al.(2024)Wen, Liang, Sierra, Luckin, Tong, Liu, Cui, and
  Tang]{wen2024ai}
Q.~Wen, J.~Liang, C.~Sierra, R.~Luckin, R.~Tong, Z.~Liu, P.~Cui, and J.~Tang.
\newblock Ai for education (ai4edu): Advancing personalized education with llm
  and adaptive learning.
\newblock In \emph{Proceedings of the 30th ACM SIGKDD Conference on Knowledge
  Discovery and Data Mining}, KDD '24, page 6743–6744, New York, NY, USA,
  2024. Association for Computing Machinery.
\newblock ISBN 9798400704901.
\newblock \doi{10.1145/3637528.3671498}.
\newblock URL \url{https://doi.org/10.1145/3637528.3671498}.

\bibitem[Whittaker et~al.(2002)Whittaker, Jones, and
  Terveen]{whittaker2002managing}
S.~Whittaker, Q.~Jones, and L.~Terveen.
\newblock Managing long term communications: conversation and contact
  management.
\newblock In \emph{Proceedings of the 35th Annual Hawaii International
  Conference on System Sciences}, pages 1070--1079, 2002.
\newblock \doi{10.1109/HICSS.2002.994063}.
\newblock URL \url{https://doi.org/10.1109/HICSS.2002.994063}.

\bibitem[Williams and Hollan(1981)]{williams1981process}
M.~D. Williams and J.~D. Hollan.
\newblock The process of retrieval from very long-term memory.
\newblock \emph{Cognitive Science}, 5\penalty0 (2):\penalty0 87--119, 1981.
\newblock ISSN 0364-0213.
\newblock URL
  \url{https://www.sciencedirect.com/science/article/pii/S0364021381800286}.

\bibitem[Williams(1992)]{williams1992simple}
R.~J. Williams.
\newblock Simple statistical gradient-following algorithms for connectionist
  reinforcement learning.
\newblock \emph{Mach. Learn.}, 8\penalty0 (3–4):\penalty0 229–256, May
  1992.
\newblock ISSN 0885-6125.
\newblock \doi{10.1007/BF00992696}.
\newblock URL \url{https://doi.org/10.1007/BF00992696}.

\bibitem[Wu et~al.(2025)Wu, Wang, Yu, Zhang, Chang, and Yu]{wu2024longmemeval}
D.~Wu, H.~Wang, W.~Yu, Y.~Zhang, K.-W. Chang, and D.~Yu.
\newblock Longmemeval: Benchmarking chat assistants on long-term interactive
  memory.
\newblock In \emph{The Thirteenth International Conference on Learning
  Representations}, 2025.
\newblock URL \url{https://openreview.net/forum?id=pZiyCaVuti}.

\bibitem[Xu et~al.(2022)Xu, Szlam, and Weston]{xu-etal-2022-beyond}
J.~Xu, A.~Szlam, and J.~Weston.
\newblock Beyond goldfish memory: Long-term open-domain conversation.
\newblock In S.~Muresan, P.~Nakov, and A.~Villavicencio, editors,
  \emph{Proceedings of the 60th Annual Meeting of the Association for
  Computational Linguistics (Volume 1: Long Papers)}, pages 5180--5197, Dublin,
  Ireland, 2022. Association for Computational Linguistics.
\newblock \doi{10.18653/v1/2022.acl-long.356}.
\newblock URL \url{https://aclanthology.org/2022.acl-long.356}.

\bibitem[Xu et~al.(2025)Xu, Mei, Gao, Tan, Liang, and Zhang]{xu2025mem}
W.~Xu, K.~Mei, H.~Gao, J.~Tan, Z.~Liang, and Y.~Zhang.
\newblock A-mem: Agentic memory for llm agents, 2025.
\newblock URL \url{https://arxiv.org/abs/2502.12110}.

\bibitem[Zhang et~al.(2024{\natexlab{a}})Zhang, Sun, Chen, Lei, Abdul-Mageed,
  Wang, Jin, Park, Yao, and Long]{zhang2024spar}
C.~Zhang, Y.~Sun, J.~Chen, J.~Lei, M.~Abdul-Mageed, S.~Wang, R.~Jin, S.~Park,
  N.~Yao, and B.~Long.
\newblock Spar: Personalized content-based recommendation via long engagement
  attention.
\newblock \emph{ArXiv preprint}, abs/2402.10555, 2024{\natexlab{a}}.
\newblock URL \url{https://arxiv.org/abs/2402.10555}.

\bibitem[Zhang et~al.(2024{\natexlab{b}})Zhang, Li, Zeng, and
  Wang]{zhang2024jasper}
D.~Zhang, J.~Li, Z.~Zeng, and F.~Wang.
\newblock Jasper and stella: distillation of sota embedding models.
\newblock \emph{ArXiv preprint}, abs/2412.19048, 2024{\natexlab{b}}.
\newblock URL \url{https://arxiv.org/abs/2412.19048}.

\bibitem[Zhang et~al.(2020)Zhang, Kishore, Wu, Weinberger, and
  Artzi]{Zhang2020BERTScore}
T.~Zhang, V.~Kishore, F.~Wu, K.~Q. Weinberger, and Y.~Artzi.
\newblock Bertscore: Evaluating text generation with {BERT}.
\newblock In \emph{8th International Conference on Learning Representations,
  {ICLR} 2020, Addis Ababa, Ethiopia, April 26-30, 2020}. OpenReview.net, 2020.
\newblock URL \url{https://openreview.net/forum?id=SkeHuCVFDr}.

\bibitem[Zhang et~al.(2024{\natexlab{c}})Zhang, Rossi, Kveton, Shao, Yang,
  Zamani, Dernoncourt, Barrow, Yu, Kim, et~al.]{zhang2024personalization}
Z.~Zhang, R.~A. Rossi, B.~Kveton, Y.~Shao, D.~Yang, H.~Zamani, F.~Dernoncourt,
  J.~Barrow, T.~Yu, S.~Kim, et~al.
\newblock Personalization of large language models: A survey.
\newblock \emph{ArXiv preprint}, abs/2411.00027, 2024{\natexlab{c}}.
\newblock URL \url{https://arxiv.org/abs/2411.00027}.

\bibitem[Zhang et~al.(2024{\natexlab{d}})Zhang, Zhang-Li, Yu, Gong, Zhou, Liu,
  Hou, and Li]{zhang2024simulating}
Z.~Zhang, D.~Zhang-Li, J.~Yu, L.~Gong, J.~Zhou, Z.~Liu, L.~Hou, and J.~Li.
\newblock Simulating classroom education with llm-empowered agents.
\newblock \emph{ArXiv preprint}, abs/2406.19226, 2024{\natexlab{d}}.
\newblock URL \url{https://arxiv.org/abs/2406.19226}.

\bibitem[Zhao et~al.(2024)Zhao, Feng, Feng, Zhong, Xu, Yang, Liu, Qin, and
  Liu]{zhao2023length}
L.~Zhao, X.~Feng, X.~Feng, W.~Zhong, D.~Xu, Q.~Yang, H.~Liu, B.~Qin, and
  T.~Liu.
\newblock Length extrapolation of transformers: A survey from the perspective
  of positional encoding.
\newblock In Y.~Al-Onaizan, M.~Bansal, and Y.-N. Chen, editors, \emph{Findings
  of the Association for Computational Linguistics: EMNLP 2024}, pages
  9959--9977, Miami, Florida, USA, Nov. 2024. Association for Computational
  Linguistics.
\newblock \doi{10.18653/v1/2024.findings-emnlp.582}.
\newblock URL \url{https://aclanthology.org/2024.findings-emnlp.582/}.

\bibitem[Zheng et~al.(2024)Zheng, Gao, Shi, Huang, Li, Xiong, Ren, Ng, Jiang,
  Li, and Li]{zheng2025dape}
C.~Zheng, Y.~Gao, H.~Shi, M.~Huang, J.~Li, J.~Xiong, X.~Ren, M.~Ng, X.~Jiang,
  Z.~Li, and Y.~Li.
\newblock Dape: Data-adaptive positional encoding for length extrapolation.
\newblock In A.~Globerson, L.~Mackey, D.~Belgrave, A.~Fan, U.~Paquet,
  J.~Tomczak, and C.~Zhang, editors, \emph{Advances in Neural Information
  Processing Systems}, volume~37, pages 26659--26700. Curran Associates, Inc.,
  2024.
\newblock URL
  \url{https://proceedings.neurips.cc/paper_files/paper/2024/file/2f050fa9f0d898e3f265d515f50ae8f9-Paper-Conference.pdf}.

\bibitem[Zhong et~al.(2024)Zhong, Guo, Gao, Ye, and Wang]{zhong2024memorybank}
W.~Zhong, L.~Guo, Q.~Gao, H.~Ye, and Y.~Wang.
\newblock Memorybank: Enhancing large language models with long-term memory.
\newblock In M.~J. Wooldridge, J.~G. Dy, and S.~Natarajan, editors,
  \emph{Thirty-Eighth {AAAI} Conference on Artificial Intelligence, {AAAI}
  2024, Thirty-Sixth Conference on Innovative Applications of Artificial
  Intelligence, {IAAI} 2024, Fourteenth Symposium on Educational Advances in
  Artificial Intelligence, {EAAI} 2014, February 20-27, 2024, Vancouver,
  Canada}, pages 19724--19731. {AAAI} Press, 2024.
\newblock \doi{10.1609/AAAI.V38I17.29946}.
\newblock URL \url{https://doi.org/10.1609/aaai.v38i17.29946}.

\end{thebibliography}

\clearpage
\appendix
\section{Implementation and Training Details}\label{app:implement}
\subsection{Parameter Setup}
We use the following hyper-parameters for all experiments:
\begin{itemize}[leftmargin=*, itemsep=1pt]
    \item \textbf{Reranker:} The reranker is an MLP with a residual connection. The training setup is:
    \begin{itemize}[leftmargin=*, itemsep=1pt]
        \item Batch size: 4
        \item Top-$M$: 5
        \item Top-$K$: 20
    \end{itemize}
    \item \textbf{Reinforcement Learning:} Retrospective Reflection uses REINFORCE with:
    \begin{itemize}[leftmargin=*, itemsep=1pt]
        \item Batch size: 4
        \item Gumbel temperature (\(\tau\)): 0.5
        \item Reward (\(R\)): \(+1\) for cited entries, \(-1\) for non-cited entries
        \item Baseline value (\(b\)): 0.5
        \item Learning rate for policy gradient updates (\(\eta\)): \(1 \times 10^{-3}\)
    \end{itemize}
    \item \textbf{LLM:} Gemini-1.5-Flash/-Pro is used for response generation with:
    \begin{itemize}[leftmargin=*, itemsep=1pt]
        \item Context window size: 128k tokens
        \item Temperature: 0.0
    \end{itemize}
    \item \textbf{Retriever:} GTE for experiments in Section~\ref{sec:offline} is pretrained with supervised contrastive learning using the following configuration:
    \begin{itemize}[leftmargin=*, itemsep=1pt]
        \item Learning rate: \(1 \times 10^{-4}\)
        \item Training epochs: 10
        \item Batch size: 32
        \item Top-$K$: 5
    \end{itemize}
\end{itemize}

\subsection{Dependencies}
Our implementation relies on the following tools and libraries:
\begin{itemize}[leftmargin=*, itemsep=1pt]
    \item \textbf{Programming Language}: Python 3.10.13
    \item \textbf{Core Libraries}: PyTorch 2.4.1+cu121, Hugging Face Transformers 4.44.2
    \item \textbf{Utilities}: NumPy, Pandas, Sklearn and Matplotlib for data processing and visualization
\end{itemize}

\subsection{Hardware and Reproducibility}
All experiments are conducted on a server with the following hardware configuration:
\begin{itemize}[leftmargin=*, itemsep=1pt]
    \item \textbf{GPUs}: 16 NVIDIA A100 GPUs
    \item \textbf{RAM}: 40 GB
    \item \textbf{CUDA Version}: 12.2
\end{itemize}

\subsection{Details for MemoryBank and LD-Agent Baselines}

\begin{wrapfigure}{r}{0.5\textwidth}
\centering
    \includegraphics[width=1\linewidth]{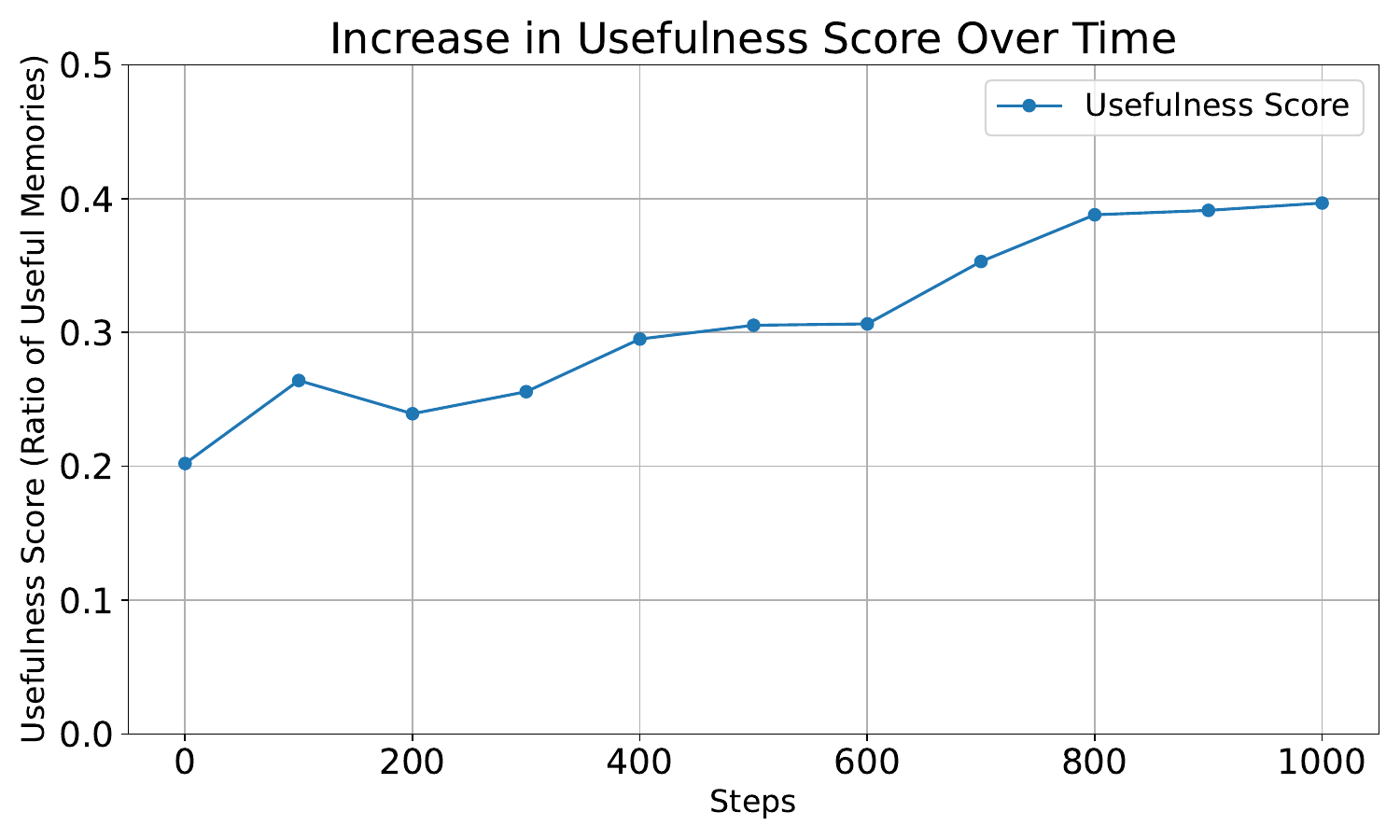}
    \caption{Convergence of usefulness scores (ratio of useful memories cited) over RL training steps. The score improves as the reranker is updated based on Retrospective Reflection, indicating enhanced alignment between retrieved memory and generated responses.}
    \label{fig:reward}
\end{wrapfigure}

We integrate MemoryBank and LD-Agent as baselines, with key features implemented using the LongMemEval codebase\footnote{\url{https://github.com/xiaowu0162/LongMemEval}}. We use Contriever as the default retriever. Particularly, they differ in the way for structuring and accessing stored information.

MemoryBank~\citep{zhong2024memorybank} retrieves historical context by maintaining a structured memory where both conversational summaries and round-level utterances are stored as key-value pairs. The retrieval process involves directly matching user queries to the most relevant stored information, ensuring efficient context retrieval for response generation.

LD-Agent~\citep{li2024hello}, on the other hand, enhances retrieval by incorporating keyphrase-based queries. In addition to storing factual and summarized information, its retrieval is based on queries with key phrases extracted from past interactions. This enables the model to adapt more effectively to diverse query formulations, retrieving context that aligns with the underlying semantic meaning of the user input.

For both methods, retrieval operates in a non-hierarchical manner, meaning that all stored data is accessed through a uniform search mechanism without additional interaction-based refinement. The retrieved content is then used to provide historical grounding for response generation. 

\subsection{The Convergence of Citation Scores in RL}\label{app:rl}

Figure~\ref{fig:reward} illustrates the convergence of citation scores (usefulness scores) during reinforcement learning. The x-axis represents the RL training steps, while the y-axis measures the ratio of useful memories cited by the LLM generator. Initially, the usefulness score starts at a low value around 0.2, reflecting the misalignment between retrieved memories and response generation. As training progresses, the score steadily increases, converging to approximately 0.4 by step 1000. This trend highlights the effectiveness of Retrospective Reflection in updating the reranker, allowing the retrieval process to better align with the generator's citation behavior. The gradual convergence indicates stable learning and suggests that RL fine-tuning improves retrieval quality without overfitting. 

\section{Dataset Description}
\label{app:dataset}
We conduct experiments on two publicly available datasets: \texttt{MSC}~\citep{xu-etal-2022-beyond} and \texttt{LongMemEval}~\citep{wu2024longmemeval}.  
{\texttt{MSC}} is a benchmark dataset for multi-session conversations, providing turn-level and session-level conversational data with annotations for relevance and response quality. On this dataset, following \citet{li2024hello}, we evaluate the ability of an LLM agent to produce human-like personalized responses. Each response can be grounded in historical context across multiple previous sessions. The focus is on accurately generating personalized responses by leveraging relevant user preferences and conversation patterns. We followed the methodology outlined by \citet{li2024hello} to construct the data for our experiments. Specifically, we use the first 1000 sessions as chat history and the rest for evaluation.

\texttt{LongMemEval} is designed for long-term conversational evaluation. It includes extended histories across turn, session, and mixed granularities. For experiments in Section~\ref{sec:offline}, we randomly sample 100 test instances and use the remaining data for training and validation.
On this dataset, following~\citet{li2024hello}, we evaluate the system's ability to answer human-designed questions about specific personal knowledge described in the historical sessions. For example, given a query like, ``What car did Mary buy last summer?'', the system must retrieve and synthesize information scattered across multiple sessions. The task emphasizes accurately identifying and leveraging relevant details from long-term memory. 

\onecolumn

\section{Case Studies}\label{app:case}

We present case studies to illustrate how RMM effectively integrates relevant memory fragments to enhance response quality. The following examples highlight scenarios where historical context is essential for maintaining coherence and accuracy in long-term dialogue.

\subsection{Case 1: Revisiting Fitness Choices (\texttt{MSC})}

Tracking personal preferences and habits across multiple conversations is essential for maintaining coherent and personalized dialogue. In this case, the user initially considers purchasing a treadmill (Session A), later expresses a preference for using the gym treadmill due to weather constraints (Session B), and finally confirms their gym-going routine (Session C). An effective memory mechanism should correctly track this evolving decision and retrieve the most up-to-date preference.

\begin{tcolorbox}[
  colframe=black!70,
  colback=white!95!gray,
  sharp corners,
  breakable,
  title=Case 1: Revisiting Fitness Choices (\texttt{MSC})
]
\small
\ttfamily

\textbf{Session A, Turn 3:}
    \begin{itemize}
        \item \textbf{Speaker\_1:} Ah, got it. Well, maybe one of the older gyms will work out better for you – or I guess you could get that \textbf{treadmill} you were talking about before.
        \item \textbf{Speaker\_2:} I'm leaning towards the \textbf{treadmill}. I think it will work better for my lifestyle.
    \end{itemize}

\textbf{Session B, Turn 2:}
    \begin{itemize}
        \item \textbf{Speaker\_1:} I go to the gym at least five times a week, and I lift weights at least three of those days. When I need to give my arms a break, I work on my leg muscles. I run around the track or just ride the stationary exercise bicycle.
        \item \textbf{Speaker\_2:} That sounds like a good plan. I definitely need to add some weights to my routine. I will be on the \textbf{treadmills} a lot, especially since it is hard for me to run outdoors daily due to the weather.
    \end{itemize}

\textbf{Session C, Turn 2:}
    \begin{itemize}
        \item \textbf{[Question] Speaker\_1:} They are great also, thanks for asking. Are you still going to the gym?
        \item \textbf{[Answer] Speaker\_2:} 
        \begin{itemize}
            \item \textbf{Ground-truth:} Yes, every night. I run on a \textbf{treadmill}.  

            \item \textbf{Output (RMM):} Yes, I go to the gym and run on the \textbf{treadmill}. It has become a key part of my routine.

            \item \textbf{Output (Long Context):} I have been considering getting a \textbf{treadmill} for home, but I am still unsure. I haven't decided yet.

        \end{itemize}
    \end{itemize}
\end{tcolorbox}

\noindent \textbf{Analysis:}  
The user’s decision about treadmill usage shifts across sessions. Initially, in Session A, they express interest in buying a treadmill. By Session B, they reconsider and decide that using the gym treadmill would be sufficient and confirm that they run on the treadmill at the gym. Without memory management, the model generates an outdated response, assuming the user is still undecided about purchasing a treadmill. 

\subsection{Case 2: Tracking Chronological Order of Events (\texttt{LongMemEval})}

In long-term interactions, correctly recalling the sequence of past events is essential for maintaining factual consistency. This case examines whether the model can track the order in which the user attended two different events.

\begin{tcolorbox}[
  colframe=black!70,
  colback=white!95!gray,
  sharp corners,
  breakable,
  title=Case 2: Tracking Chronological Order of Events (\texttt{LongMemEval})
]
\small
\ttfamily
\textbf{Session A, Turn 1:}
    \begin{itemize}
        \item \textbf{Speaker\_1:} I recently attended a workshop on \textbf{Effective Time Management} at the local community center. It was incredibly insightful and gave me some strategies for managing tasks efficiently. What are your thoughts on time management workshops?
        \item \textbf{Speaker\_2:} They're quite helpful for understanding basic techniques, but it also depends on how you apply the strategies in daily life.
    \end{itemize}

\textbf{Session B, Turn 3:}
    \begin{itemize}
        \item \textbf{Speaker\_1:} I also attended a webinar on \textbf{Data Analysis using Python} two months ago. The instructor shared some incredible resources for data visualization. Have you explored advanced visualization techniques?
        \item \textbf{Speaker\_2:} Yes, they can greatly enhance how you present your data. Libraries like \textbf{Matplotlib} and \textbf{Seaborn} are good starting points for creating professional visuals.
    \end{itemize}

\vspace{1mm}
\textbf{[Question]} Which event did I attend first, the ``Effective Time Management'' workshop or the ``Data Analysis using Python'' webinar?

\vspace{1mm}
\textbf{[Answer]}
\begin{itemize}
    \item \textbf{Ground-truth:} ``Data Analysis using Python'' webinar.
    \item \textbf{Output (RMM):} You attended the \textbf{Data Analysis using Python} webinar two months ago. The \textbf{Effective Time Management} workshop happened later at the local community center.
    \item \textbf{Output (Long Context):} I'm not sure, but you mentioned both events in previous conversations.
\end{itemize}
\end{tcolorbox}

\noindent \textbf{Analysis:} The correct response requires linking the time reference (``two months ago'') with the corresponding event. Without RMM, the model fails to retrieve this detail, resulting in an uncertain and incomplete answer. With RMM, the model correctly recalls the chronological order, demonstrating the advantage of structured memory retrieval in tracking event sequences.

\section{Prompts}\label{app:prompt}
\subsection{Prospective Reflection}
\subsubsection{Memory Extraction}\label{app:p2}

\textbf{Function: } Memory extraction for \texttt{SPEAKER\_1}
\begin{tcolorbox}[breakable]
\small
\ttfamily

\textbf{Task Description:} Given a session of dialogue between SPEAKER\_1 and SPEAKER\_2, extract the personal summaries of SPEAKER\_1, with references to the corresponding turn IDs. Ensure the output adheres to the following rules:
\begin{itemize}
    \item Output results in \textbf{JSON format}. The top-level key is ``extracted\_memories''. The value should be a list of dictionaries, where each dictionary has the keys ``summary'' and ``reference'':
    \begin{itemize}
        \item \textbf{summary}: A concise personal summary, which captures relevant information about SPEAKER\_1's experiences, preferences, and background, across multiple turns.
        \item \textbf{reference}: A list of references, each in the format of \texttt{[turn\_id]} indicating where the information appears.
    \end{itemize}
    \item If no personal summary can be extracted, return \texttt{NO\_TRAIT}.
\end{itemize}

\vspace{2mm}
\textbf{Example:}

\textcolor{black}{\textbf{INPUT:}}
\begin{itemize}
    \item \textbf{Turn 0:}
    \begin{itemize}
        \item \textbf{SPEAKER\_1:} Did you check out that new gym in town?
        \item \textbf{SPEAKER\_2:} Yeah, I did. I'm not sure I like the vibe there, though.
    \end{itemize}

    \item \textbf{Turn 1:}
    \begin{itemize}
        \item \textbf{SPEAKER\_1:} What was wrong with it?
        \item \textbf{SPEAKER\_2:} The folks there seemed to care more about how they looked than working out. It was a little too trendy for me. I'm pretty plain.
    \end{itemize}

    \item \textbf{Turn 2:}
    \begin{itemize}
        \item \textbf{SPEAKER\_1:} Ah, got it. Well, maybe one of the older gyms will work out better for you—or I guess you could get that treadmill you were talking about before. Are you leaning one way or the other yet?
        \item \textbf{SPEAKER\_2:} I'm leaning towards the treadmill. I think it will work better for my lifestyle. I just don't know which type to get. There are so many choices out there. Do you use a treadmill at your gym? Do you have a suggestion for a home one?
    \end{itemize}

    \item \textbf{Turn 3:}
    \begin{itemize}
        \item \textbf{SPEAKER\_1:} I usually just lift weights there, to be honest. But I think I've heard good things about the NordicTrack?
        \item \textbf{SPEAKER\_2:} Yeah, I've heard good things about that, too. I like the idea of a multi-exercise piece of equipment. As long as the weather isn't too bad, then I prefer to go for a run. But since it rains quite a bit here, I like the idea of an inside option. How is the weather in New England?
    \end{itemize}

    \item \textbf{Turn 4:}
    \begin{itemize}
        \item \textbf{SPEAKER\_1:} Oh, it can get pretty foggy and rainy here too, I'm afraid. But as I'm sure you've heard, it's really beautiful in the fall! Are there four distinct seasons where you are, too?
        \item \textbf{SPEAKER\_2:} Yes, I've heard about the fall colors. I may get there one day. Yes, we have seasons—rain, lighter rain, summer, and more rain! Ha!
    \end{itemize}

    \item \textbf{Turn 5:}
    \begin{itemize}
        \item \textbf{SPEAKER\_1:} Haha! I lived overseas in the tropics once. Sounds just like it!
        \item \textbf{SPEAKER\_2:} The tropics sound great. It's not as warm as the tropics, but I like it. I'm from Alaska, so I'm pretty weather-tough.
    \end{itemize}
\end{itemize}

\textcolor{black}{\textbf{OUTPUT:}}
\begin{lstlisting}[backgroundcolor=\color{gray!10}, frame=none]
{
    "extracted_memories": [
        {
            "summary": "SPEAKER_1 asked about a new gym in town and suggested older gyms or a treadmill as alternatives.",
            "reference": [0, 2]
        },
        {
            "summary": "SPEAKER_1 usually lifts weights at the gym rather than using a treadmill.",
            "reference": [3]
        },
        {
            "summary": "SPEAKER_1 has heard good things about the NordicTrack treadmill.",
            "reference": [3]
        },
        {
            "summary": "SPEAKER_1 lives in New England and experiences foggy and rainy weather but enjoys the fall season.",
            "reference": [4]
        },
        {
            "summary": "SPEAKER_1 has lived overseas in the tropics before.",
            "reference": [5]
        }
    ]
}
\end{lstlisting}

\textbf{Task:} Follow the JSON format demonstrated in the example above and extract the personal summaries for SPEAKER\_1 from the following dialogue session.

\textbf{Input:} \{\}

\textbf{Output:}
\end{tcolorbox}

\textbf{Function: } Memory extraction for \texttt{SPEAKER\_2}
\begin{tcolorbox}[breakable]
\small
\ttfamily

\textbf{Task Description:} Given a session of dialogue between SPEAKER\_1 and SPEAKER\_2, extract the personal summaries of SPEAKER\_2, with references to the corresponding turn IDs. Ensure the output adheres to the following rules:
\begin{itemize}
    \item Output results in \textbf{JSON format}. The top-level key is ``extracted\_memories''. The value should be a list of dictionaries, where each dictionary has the keys ``summary'' and ``reference'':
    \begin{itemize}
        \item \textbf{summary}: A concise personal summary, which captures relevant information about SPEAKER\_2's experiences, preferences, and background, across multiple turns.
        \item \textbf{reference}: A list of references, each in the format of \texttt{[turn\_id]} indicating where the information appears.
    \end{itemize}
    \item If no personal summary can be extracted, return \texttt{NO\_TRAIT}.
\end{itemize}

\vspace{2mm}
\textbf{Example:}

\textcolor{black}{\textbf{INPUT:}}
\begin{itemize}
    \item \textbf{Turn 0:}
    \begin{itemize}
        \item \textbf{SPEAKER\_1:} Did you manage to go out on a run today?
        \item \textbf{SPEAKER\_2:} Yes, I actually was able to. I am considering joining the local gym. Do you prefer going to the gym?
    \end{itemize}

    \item \textbf{Turn 1:}
    \begin{itemize}
        \item \textbf{SPEAKER\_1:} I do actually. I like the controlled environment. I don't want to have to depend on the weather considering where I live.
        \item \textbf{SPEAKER\_2:} That's why I am thinking about it. I hate to have to run when it's raining, and I feel like it rains here all the time.
    \end{itemize}

    \item \textbf{Turn 2:}
    \begin{itemize}
        \item \textbf{SPEAKER\_1:} A lot of gyms have tracks so that you can run indoors. Hey, have you thought about maybe buying a treadmill and using that at home?
        \item \textbf{SPEAKER\_2:} I am definitely considering getting one. I'm just trying to figure out what I would do more—go to the gym and actually do more than just running, or stick to what I know and get a treadmill.
    \end{itemize}

    \item \textbf{Turn 3:}
    \begin{itemize}
        \item \textbf{SPEAKER\_1:} Oh, that's true. I hadn't thought about all of that. You're right. With a gym, there are a whole lot of options for what you can do. Do you have some good gyms near you?
        \item \textbf{SPEAKER\_2:} They just built one in the small town really close to me, and it looks pretty decent. Before that, it was like an hour drive.
    \end{itemize}

    \item \textbf{Turn 4:}
    \begin{itemize}
        \item \textbf{SPEAKER\_1:} With you not owning a car, going to any others would probably be difficult. Well, do you have any good parks and running trails nearby?
        \item \textbf{SPEAKER\_2:} Yeah, exactly. There is a super nice little running trail that is pretty decent.
    \end{itemize}

    \item \textbf{Turn 5:}
    \begin{itemize}
        \item \textbf{SPEAKER\_1:} Hey, do you run with anyone? I mean, have you joined a club, or will you if you haven't?
        \item \textbf{SPEAKER\_2:} There isn’t any around here; maybe I could start one. Thank you for that idea.
    \end{itemize}
\end{itemize}

\textcolor{black}{\textbf{OUTPUT:}}
\begin{lstlisting}[backgroundcolor=\color{gray!10}, frame=none]
{
    "extracted_memories": [
        {
            "summary": "SPEAKER_2 is considering joining a local gym due to frequent rain affecting outdoor runs.",
            "reference": [0, 1]
        },
        {
            "summary": "SPEAKER_2 is debating between buying a treadmill for home use or going to the gym for more workout variety.",
            "reference": [2]
        },
        {
            "summary": "A new gym was recently built nearby SPEAKER_2, replacing a previous one that was an hour away.",
            "reference": [3]
        },
        {
            "summary": "SPEAKER_2 has access to a nice local running trail.",
            "reference": [4]
        },
        {
            "summary": "SPEAKER_2 notices there is no local running club but is considering starting one.",
            "reference": [5]
        }
    ]
}
\end{lstlisting}

\textbf{Task:} Follow the JSON format demonstrated in the example above and extract the personal summaries for SPEAKER\_2 from the following dialogue session.

\textbf{Input:} \{\}

\textbf{Output:}
\end{tcolorbox}

\newpage
\subsubsection{Memory Update}\label{app:p3}
\begin{tcolorbox}[breakable]
\small
\ttfamily

\textbf{Task Description:} Given a list of history personal summaries for a specific user and a new and similar personal summary from the same user, update the personal history summaries following the instructions below:

\begin{itemize}
    \item \textbf{Input format:} Both the history personal summaries and the new personal summary are provided in JSON format, with the top-level keys of ``history\_summaries'' and ``new\_summary''.
    \item \textbf{Possible update actions:}
    \begin{itemize}
        \item \textbf{Add:} If the new personal summary is not relevant to any history personal summary, add it. \\
        \textbf{Format:} \texttt{Add()}
        \item \textbf{Merge:} If the new personal summary is relevant to a history personal summary, merge them as an updated summary. \\
        \textbf{Format:} \texttt{Merge(index, merged\_summary)} \\
        \textbf{Note:} \texttt{index} is the position of the relevant history summary in the list. \texttt{merged\_summary} is the merged summary of the new summary and the relevant history summary. Two summaries are considered relevant if they discuss the same aspect of the user's personal information or experiences.
    \end{itemize}
    \item If multiple actions need to be executed, output each action in a single line, and separate them with a newline character (\texttt{"\textbackslash n"}).
    \item Do not include additional explanations or examples in the output—only return the required action functions.
\end{itemize}

\vspace{2mm}
\textbf{Example:}

\textcolor{black}{\textbf{INPUT:}}
\begin{itemize}
    \item \textbf{History Personal Summaries:}
    \begin{itemize}
        \item \texttt{\{"history\_summaries": ["SPEAKER\_1 works out although he doesn't particularly enjoy it."]\}}
    \end{itemize}
    \item \textbf{New Personal Summary:}
    \begin{itemize}
        \item \texttt{\{"new\_summary": "SPEAKER\_1 exercises every Monday and Thursday."\}}
    \end{itemize}
\end{itemize}

\textcolor{black}{\textbf{OUTPUT ACTION:}}

\texttt{Merge(0, SPEAKER\_1 exercises every Monday and Thursday, although he doesn't particularly enjoy it.)}

\vspace{2mm}
\textbf{Task:} Follow the example format above to update the personal history for the given case.

\textcolor{black}{\textbf{INPUT:}}
\begin{itemize}
    \item \textbf{History Personal Summaries:} \texttt{\{\}}
    \item \textbf{New Personal Summary:} \texttt{\{\}}
\end{itemize}

\textcolor{black}{\textbf{OUTPUT ACTION:}}
\end{tcolorbox}

\newpage
\subsection{Retrospective Reflection}\label{app:p1}
\begin{tcolorbox}[breakable]
\small
\ttfamily

\textbf{Task Description:} Given a user query and a list of memories consisting of personal summaries with their corresponding original turns, generate a natural and fluent response while adhering to the following guidelines:
\begin{itemize}
    \item Cite \textbf{useful} memories using \([i]\), where \(i\) corresponds to the index of the cited memory.
    \item Do \textbf{not cite memories that are not useful}. If no useful memory exist, output \texttt{[NO\_CITE]}.
    \item Each memory is independent and may repeat or contradict others. The response must be directly supported by cited memories.
    \item If the response relies on multiple memories, list all corresponding indices, e.g., \([i, j, k]\).
    \item The citation is evaluated based on whether the response references the original turns, \textbf{not the summaries}.
\end{itemize}

\vspace{2mm}
\textbf{Examples:}

\textcolor{black}{\textbf{Case 1: Useful Memories Found}}

\textcolor{black}{\textbf{INPUT:}}
\begin{itemize}
    \item \textbf{User Query:} SPEAKER\_1: What hobbies do I enjoy?
    \item \textbf{Memories:}
    \begin{itemize}
        \item \textbf{Memory [0]:} SPEAKER\_1 enjoys hiking and often goes on weekend trips.
        \begin{itemize}
            \item[*] \texttt{Speaker 1: I love spending my weekends hiking in the mountains.} \\
                     \texttt{Speaker 2: That sounds amazing! Do you go alone or with friends?}
            \item[*] \texttt{Speaker 1: Last month, I hiked a new trail and it was amazing.} \\
                     \texttt{Speaker 2: Nice! Which trail was it?}
        \end{itemize}
        \item \textbf{Memory [1]:} SPEAKER\_1 plays the guitar and occasionally performs at open mics.
        \begin{itemize}
            \item[*] \texttt{Speaker 1: I’ve been practicing guitar for years and love playing at open mics.} \\
                     \texttt{Speaker 2: That’s awesome! What songs do you usually play?}
            \item[*] \texttt{Speaker 1: I performed at a local cafe last week and had a great time.} \\
                     \texttt{Speaker 2: That must have been fun! Were there a lot of people?}
        \end{itemize}
        \item \textbf{Memory [2]:} SPEAKER\_1 is interested in astronomy and enjoys stargazing.
        \begin{itemize}
            \item[*] \texttt{Speaker 1: I recently bought a telescope to get a closer look at planets.} \\
                     \texttt{Speaker 2: That’s so cool! What have you seen so far?}
            \item[*] \texttt{Speaker 1: I love stargazing, especially when there’s a meteor shower.} \\
                     \texttt{Speaker 2: I’d love to do that sometime. When’s the next one?}
        \end{itemize}
    \end{itemize}
\end{itemize}

\textbf{Output:} You enjoy hiking, playing the guitar, and stargazing. [0, 1, 2]

\vspace{3mm}

\textcolor{black}{\textbf{{Case 2: No Useful Memories}}}

\textcolor{black}{\textbf{INPUT:}}
\begin{itemize}
    \item \textbf{User Query:} SPEAKER\_1: What countries did I go to last summer?
    \item \textbf{Memories:}
    \begin{itemize}
        \item \textbf{Memory [0]:} SPEAKER\_1 enjoys hiking and often goes on weekend trips.
        \begin{itemize}
            \item[*] \texttt{Speaker 1: I love spending my weekends hiking in the mountains.} \\
                     \texttt{Speaker 2: That sounds amazing! Do you go alone or with friends?}
            \item[*] \texttt{Speaker 1: Last month, I hiked a new trail and it was amazing.} \\
                     \texttt{Speaker 2: Nice! Which trail was it?}
        \end{itemize}
        \item \textbf{Memory [1]:} SPEAKER\_1 plays the guitar and occasionally performs at open mics.
        \begin{itemize}
            \item[*] \texttt{Speaker 1: I’ve been practicing guitar for years and love playing at open mics.} \\
                     \texttt{Speaker 2: That’s awesome! What songs do you usually play?}
            \item[*] \texttt{Speaker 1: I performed at a local cafe last week and had a great time.} \\
                     \texttt{Speaker 2: That must have been fun! Were there a lot of people?}
        \end{itemize}
        \item \textbf{Memory [2]:} SPEAKER\_1 is interested in astronomy and enjoys stargazing.
        \begin{itemize}
            \item[*] \texttt{Speaker 1: I recently bought a telescope to get a closer look at planets.} \\
                     \texttt{Speaker 2: That’s so cool! What have you seen so far?}
            \item[*] \texttt{Speaker 1: I love stargazing, especially when there’s a meteor shower.} \\
                     \texttt{Speaker 2: I’d love to do that sometime. When’s the next one?}
        \end{itemize}
    \end{itemize}
\end{itemize}

\textbf{Output:} I don't have enough information to answer that. [NO\_CITE]

\vspace{2mm}
\textbf{Additional Instructions:}
\begin{itemize}
    \item Ensure the response is fluent and directly answers the user's query.
    \item Always cite the useful memory indices explicitly.
    \item The citation is evaluated based on whether the response references the original turns, \textbf{not the summaries}.
    \item Follow the format of the examples provided above.
\end{itemize}

\vspace{2mm}
\textbf{Input:}
\begin{itemize}
    \item \textbf{User Query:} \{\}
    \item \textbf{Memories:} \{\}
\end{itemize}

\textcolor{black}{\textbf{Output:}} 

\end{tcolorbox}

\subsection{LLM-as-a-Judge}\label{app:judge}
\begin{tcolorbox}[breakable]
\small
\ttfamily

You are an expert language model evaluator. I will provide you with a question, a ground-truth answer, and a model-generated response. Your task is to determine whether the response correctly answers the question by following these evaluation rules:
\begin{itemize}
    \item Answer \textbf{Yes} if the response contains or directly matches the correct answer.
    \item Answer \textbf{Yes} if the response includes all necessary intermediate steps leading to the correct answer.
    \item Answer \textbf{No} if the response provides only a partial answer or omits essential information.
    \item Answer \textbf{No} if the response does not sufficiently address the question.
\end{itemize}

\vspace{2mm}
\textbf{Examples:}

\textcolor{black}{\textbf{Example 1: Correct Response}}
\begin{itemize}
    \item \textbf{Question:} What is the capital of France?
    \item \textbf{Ground-truth Answer:} Paris
    \item \textbf{Response:} The capital of France is Paris.
\end{itemize}

\textcolor{black}{\textbf{Evaluation:}}
\begin{itemize}
    \item \textbf{Output:} Yes
\end{itemize}

\vspace{2mm}
\textcolor{black}{\textbf{Example 2: Incorrect Response}}
\begin{itemize}
    \item \textbf{Question:} What is the capital of France?
    \item \textbf{Ground-truth Answer:} Paris
    \item \textbf{Response:} France is a country in Europe.
\end{itemize}

\textcolor{black}{\textbf{Evaluation:}}
\begin{itemize}
    \item \textbf{Output:} No
\end{itemize}

\vspace{2mm}
\textbf{Additional Instructions:}
\begin{itemize}
    \item Apply the evaluation criteria consistently.
    \item Base your decision strictly on the information in the response.
    \item Avoid subjective interpretations and adhere to the provided examples.
\end{itemize}

\vspace{2mm}
\textbf{Input:}
\begin{itemize}
    \item \textbf{Question:} \{\}
    \item \textbf{Ground-truth Answer:} \{\}
    \item \textbf{Response:} \{\}
\end{itemize}

\textcolor{black}{\textbf{Output:}}

\end{tcolorbox}

\end{document}